\documentclass[10pt,twocolumn,letterpaper]{article}
\usepackage[table]{xcolor}
\usepackage{graphicx}  % For including graphics
\usepackage{amsmath}   % For advanced math features
\usepackage{amsfonts}  % For additional fonts (like \mathbb)
\usepackage{caption}   % For customizing captions
\usepackage{booktabs} % For better horizontal lines
\usepackage{multirow} % For multi-row cells
\usepackage{array}    % For fixed-width columns
\usepackage{microtype}  % Better typography
\usepackage{bbm}
\usepackage{adjustbox}
\usepackage{float}
\usepackage{svg}
\usepackage[accsupp]{axessibility}  % Improves PDF readability for those with disabilities.
\usepackage[pagenumbers]{cvpr} % To force page numbers, e.g. for an arXiv version
\newcommand{\REF}[1]{{\color{red}REF}}

\newcommand{\Fig}[1]{\hyperref[{#1}]{Figure~\ref*{#1}}}  % beginning of sentence
\newcommand{\fig}[1]{\hyperref[{#1}]{Fig.~\ref*{#1}}}    % somewhere
\newcommand{\Tab}[1]{\hyperref[{#1}]{Table~\ref*{#1}}}
\newcommand{\tab}[1]{\hyperref[{#1}]{Table~\ref*{#1}}}
\newcommand{\Eqn}[1]{\hyperref[{#1}]{Equation~\ref*{#1}}}
\newcommand{\eqn}[1]{\hyperref[{#1}]{Eq.~\ref*{#1}}} % equation 1.1
\newcommand{\sect}[1]{\hyperref[{#1}]{Sec.~\ref*{#1}}} % Section 1
\newcommand{\supp}[1]{\hyperref[{#1}]{Suppl.~\ref*{#1}}}
\newcommand{\app}[1]{\hyperref[{#1}]{App.~\ref*{#1}}}
\newcommand{\App}[1]{\hyperref[{#1}]{Appendix~\ref*{#1}}}

\newcommand{\method}{\textsc{Slot Contrast}\xspace}
\newcommand{\loss}{slot-slot contrastive loss\xspace}

\newcommand{\RTab}[1]{\hyperref[{#1}]{Tab.~R\ref*{#1}}}
\newcommand{\Rtab}[1]{\hyperref[{#1}]{Tab.~R\ref*{#1}}}

\newcommand{\Rfig}[1]{\hyperref[{#1}]{ Fig.~R\ref*{#1}}}

\makeatletter
\newcommand*{\addFileDependency}[1]{% argument=file name and extension
  \typeout{(#1)}
  \@addtofilelist{#1}
  \IfFileExists{#1}{}{\typeout{No file #1.}}
}
\makeatother
\newcommand*{\myexternaldocument}[1]{%
    \externaldocument{#1}%
    \addFileDependency{#1.tex}%
    \addFileDependency{#1.aux}%
  }

\definecolor{ourblue}{RGB}{9,134,223} % 0986dfff
\definecolor{ourdarkblue}{RGB}{8,113,185} % 0871b9ff

\definecolor{ourorange}{RGB}{224,90,18} % e05a12ff
\definecolor{ourdarkorange}{RGB}{191,77,16} % bf4d10ff
\definecolor{ourlightorange}{RGB}{241,207,184} % f1cfb8ff

\definecolor{ouryellow}{RGB}{227,213,25} % e3d519ff
\definecolor{ourdarkyellow}{RGB}{197,185,0} % c5b900ff

\definecolor{ourpink}{RGB}{247,24,139} % f7188bff
\definecolor{ourdarkpink}{RGB}{198,20,112} % c61470ff
\definecolor{ourlightpink}{RGB}{247,210,229} % f7d2e5ff

\definecolor{ourgreen}{RGB}{159,198,52} % 9fc634ff
\definecolor{ourdarkgreen}{RGB}{118,146,39} % 769227ff

\colorlet{TableColor}{ourlightorange}
\usepackage[pagebackref]{hyperref}
\usepackage{url}
\hypersetup{colorlinks,
            linkcolor=ourdarkblue,
            urlcolor=ourdarkblue,
            citecolor=ourdarkblue}

\definecolor{cvprblue}{rgb}{0.21,0.49,0.74}
\def\paperID{16593} % *** Enter the Paper ID here
\def\confName{CVPR}
\def\confYear{2025}

\usepackage{xr-hyper}
\myexternaldocument{suppl}

\title{Temporally Consistent Object-Centric Learning by Contrasting Slots}
\author{
Anna Manasyan\textsuperscript{1} \quad Maximilian Seitzer\textsuperscript{1} \quad Filip Radovic\textsuperscript{1} \quad Georg Martius\textsuperscript{1,3} \quad Andrii Zadaianchuk\textsuperscript{2} \\
\textsuperscript{1}University of Tübingen, Tübingen, Germany \quad
\textsuperscript{2}University of Amsterdam, Amsterdam, Netherlands \\
\textsuperscript{3}Max Planck Institute for Intelligent Systems, Tübingen, Germany \\
{\tt\small anna.manasyan@student.uni-tuebingen.de \quad a.zadaianchuk@uva.nl}
}

\begin{document}
%------------------------------------------------------------------------------
% place teaser between title and abstract
\makeatletter
\apptocmd\@maketitle{{\teaser{}}}{}{}
\makeatother
%------------------------------------------------------------------------------

\newcommand{\teaser}{%
%------------------------------------------------------------------------------
\vspace{-6pt}
\centering
\setlength{\tabcolsep}{1pt}

\vspace{-6pt}
    \centering
    \includegraphics[width=\textwidth]{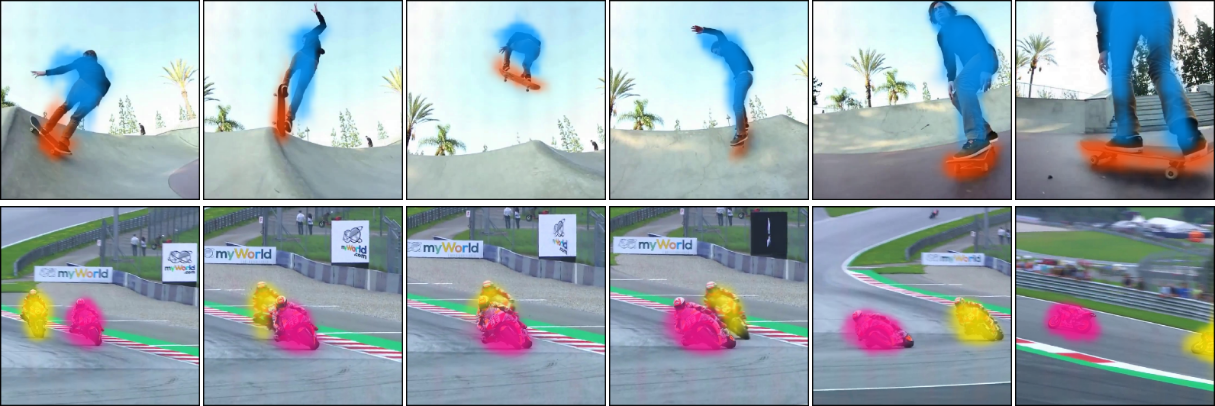}%
    \captionof{figure}{\method: Our method introduces a temporal contrastive loss that enhances temporal consistency in unsupervised video object-centric models. It stabilizes representations across frames, achieving state-of-the-art object discovery on complex real-world videos.\looseness=-1}
    \label{fig:teaser}
%------------------------------------------------------------------------------
\par\vspace{2em}
%------------------------------------------------------------------------------
}
\maketitle
\begin{abstract}
Unsupervised object-centric learning from videos is a promising approach to extract structured representations from large, unlabeled collections of videos. 
To support downstream tasks like autonomous control, these representations must be both compositional and temporally consistent. 
Existing approaches based on recurrent processing often lack long-term stability across frames because their training objective does not enforce temporal consistency.
In this work, we introduce a novel object-level temporal contrastive loss for video object-centric models that explicitly promotes temporal consistency.
Our method significantly improves the temporal consistency of the learned object-centric representations, yielding more reliable video decompositions that facilitate challenging downstream tasks such as unsupervised object dynamics prediction.
Furthermore, the inductive bias added by our loss strongly improves object discovery, leading to state-of-the-art results on both synthetic and real-world datasets, outperforming even weakly-supervised methods that leverage motion masks as additional cues. 
%\\ The website can be found at \href{https://slotcontrast.github.io/}{slotcontrast.github.io}.
\\ Visit \href{https://slotcontrast.github.io/}{slotcontrast.github.io} for videos and further details.

\end{abstract}

\section{Introduction}
\label{sec:intro}

Object-centric learning (OCL)~\citep{burgess2019monet, greff2019multi, Locatello2020SlotAttention, seitzer2023bridging} is a rapidly advancing area of visual representation learning that enables autonomous systems to represent, understand, and model high-dimensional data directly in terms of its constituent entities. 
Structured object-centric representations (often referred to as \emph{slots}~\citep{Locatello2020SlotAttention}) facilitate generalization and robustness~\citep{dittadi2021generalization, didolkar2024zero} of scene representations across diverse downstream tasks, from visual question answering~\citep{VQA,mamaghan2024exploring,Xu2024slotvlm, didolkar2025ctrlo} to control~\citep{zadaianchuk2020self, driess2023palme, yoon2023investigation, haramati2024entitycentric}. 
Of particular interest are video-based object-centric methods%
% Particularly interesting, video-based object-centric methods
~\cite{kipf2022conditional, Jiang2020SCALOR, elsayed2022savi++, Singh2022STEVE, aydemir2023self, wu2023slotdiffusion, zadaianchuk2023objectcentric} that learn to represent objects that evolve and interact over time. 
These representations make the methods powerful tools for applications such as unsupervised online object tracking~\citep{wang2019unsupervised, meng2023tracking} and structured world modeling~\citep{kipf2019contrastive, veerapaneni2019entity, wu2023slotformer}.
% \begin{figure}[tbh]
%     \centering
%     \rule{8cm}{5cm} % This is the placeholder
%     \caption{Temporal Consistency Problem \Andr{add figure with problem visualization}.}
%     \label{fig:consitency_problem}
% \end{figure}
Unsupervised object-centric learning on videos has seen significant progress in recent years~\citep{aydemir2023self, wu2023slotdiffusion,zadaianchuk2023objectcentric}, mainly due to the use of pre-trained representations from self-supervised foundational models~\citep{Caron2021DINO, oquab2023dinov2} coupled with diverse training datasets like YouTube-VIS~\citep{vis2019, vis2021}. 
Nevertheless, these methods still face significant challenges, especially maintaining consistent object-centric representations across time and uniquely representing each object---critical factors for successful multi-object tracking and modeling of dynamic scenes~\citep{kipf2022conditional, veerapaneni2019entity, wu2023slotformer}.

% Temporal consistency~\citep{yu2024vonet, li2024reasoning} in object-centric representations refers to assigning the same representation slot to the same object throughout a video sequence, effectively maintaining an object-specific identifier over time. 
Temporal consistency~\citep{greff2020binding, yu2024vonet, li2024reasoning} in object-centric representations refers to maintaining the same representation placeholder, called slot, for an object throughout a video sequence, effectively serving as a stable object-specific identifier over time.
% Initially, temporal consistency was tackled in supervised object tracking~\cite{bewley2016simple}, which requires a lot of human annotations and is limited to domains with robust object detectors. 
Existing unsupervised object-centric methods~\citep{kipf2019contrastive, traub2023looping, meo2024object} aiming to discover consistent representations have primarily been studied on toy datasets with limited complexity~\citep{johnson2017clevr, Karazija2021clevrtex, tangemann2021unsupervised}. 
In contrast, real-world video sequences present numerous challenges, including object occlusions, reappearances, and complex multi-object interactions, which complicate maintaining consistent object representations.

In this paper, we introduce a novel method to address the challenge of maintaining consistent temporal representations in object-centric models, extending the line of research on slot-based unsupervised video models~\citep{zadaianchuk2023objectcentric,elsayed2022savi++}. 
Our approach (named \method) scales to real-world video data and produces consistent object-centric representations. 
Notably, it achieves these results without requiring any human annotations.
% In this paper, we propose a novel method to address the temporal representation consistency challenge that scales to real-world video data while producing consistent object-centric representations without any human annotations. 
In particular, we propose a novel self-supervised contrastive learning objective, 
which contrasts slot representations throughout the batch while ensuring temporal coherence across consecutive frames. 
In addition, we modify the slot's initialization strategy~\citep{Locatello2020SlotAttention} to promote distinct, contrastive representations. 
%To do so effectively, we modify the slot's initialization strategy~\citep{Locatello2020SlotAttention} to promote distinct, contrastive representations. 
This combination leads to improved temporal consistency of learned representations, which we show to be highly effective for challenging downstream tasks such as unsupervised object tracking and latent object dynamics learning. 

Overall, our contributions are as follows:

\begin{itemize}
\item We propose the novel \loss that sets the state-of-the-art in temporal consistency when integrated into slot-based video processing methods. 
\item We develop \method, a simple and effective OCL architecture using the \loss paired with learned initialization that scales to real-world data, such as YouTube videos. 
%\item We develop \method, a simple and complete architecture a simple and effective OCL method that learns consistent object-centric representations across entire video sequences using a novel slot-slot contrastive objective and scales to real-world data like YouTube videos. 
\item We extensively study the usefulness of our learned object-centric representations for challenging downstream tasks, including unsupervised online tracking with complete occlusions and latent object dynamics modeling.
\item We show that \method does not only improve the temporal consistency of the representations, but also achieves state-of-the-art on the object discovery task, outperforming weakly-supervised models using motion cues. 

% \item We propose an \emph{OcclusionMOVi} dataset and corresponding evaluation that enables a controlled study of the video object-centric representations consistency in complex video scenarios involving full occlusions and object reappearances.

\end{itemize}

\section{Related Work}
\label{sec:related_work}

\paragraph{Unsupervised video object-centric learning}
There exists an extensive body of research~\citep{Kosiorek2018SQAIR, Jiang2020SCALOR,  vanstennkiste2018relational,veerapaneni2019entity,greff2019multi,kabra2021simone,kipf2022conditional,elsayed2022savi++,Singh2022STEVE,traub2023learning,safadoust2023multi, aydemir2023self, zadaianchuk2023objectcentric} on discovering objects from video without any human annotations, primarily through tracking either object bounding boxes or masks. 
To achieve this, most of these works combine an auto-encoder framework with a simple reconstruction objective, adding inductive biases for object discovery through structured encoders~\citep{burgess2019monet, Locatello2020SlotAttention} and decoders~\citep{watters2019spatial}. 
In particular, many modern object-centric image models~\citep{seitzer2023bridging,  wu2023slotdiffusion, jiang2023object, kakogeorgiou2024spot, didolkar2025ctrlo} use a latent slot attention module~\citep{Locatello2020SlotAttention} to extract object representations and corresponding object masks. 
For video data, most current methods~\citep{kipf2022conditional, Zoran2021parts, elsayed2022savi++, Singh2022STEVE, traub2023learning, zadaianchuk2023objectcentric} connect slots across frames, with slots from the previous frame initializing those in the current frame. 
Notably, recent approaches~\cite{zadaianchuk2023objectcentric, qian2023semantics, aydemir2023self} have successfully scaled object discovery to real-world unconstrained videos. 
To achieve this, SOLV~\citep{aydemir2023self} introduces temporal consistency via agglomerative clustering and prediction of middle-frame features, whereas VideoSAUR~\citep{zadaianchuk2023objectcentric} learns object-centric representations by predicting temporal similarity of self-supervised features~\citep{Caron2021DINO,oquab2023dinov2}. 
While such methods can decompose short videos, they still struggle with long-term temporal consistency. 
In contrast, we show that learning representations that are both informative and contrastive can significantly enhance both object discovery and temporal consistency on longer videos.\looseness=-1

% Slot Attention (SA) ~\citep{Locatello2020SlotAttention} has recently become a foundational approach in object-centric learning, with extensions developed to handle video data ~\cite{kipf2022conditional, elsayed2022savi++, singh2022simple}. To adapt SA for real-world applications, researchers have also incorporated self-supervised DINO models ~\cite{caron2021emerging, oquab2023dinov2}, effectively scaling SA to complex, real-world visual data. ~\cite{seitzer2023bridging, zadaianchuk2023objectcentric, aydemir2023self}. The predominate architecture for video slot attention is RNN-based which sometimes can have poor stability and capacity therefore ~\citet{singh2024parallelized} proposes a temporally-parallelization slot learning architecture. While substantial progress has been made in unsupervised object-centric learning for sequential data, maintaining temporal consistency remains a significant challenge. This capability is essential for the representations to be useful in real-world applications. 

\paragraph{Temporal Consistency} Achieving temporal consistency is essential for any computer vision task involving video data, whether it is tracking points, bounding boxes, segmentation masks, optical flow, or representations~\citep{lu2024self,tokmakov2022object,veerapaneni2019entity,wang2019learning,van2023tracking,kirillov2023segment, ravi2024sam2}.
% In SA-based video object-centric learning, we need temporal consistency over slots.  
% Current unsupervised video
% decomposition methods stay consistent in time, initializing the object discovery process with the previous time step representations. 
% However, this approach is not always enough, especially in highly dynamic scenes. 
In object-centric learning for videos, a range of different approaches have been proposed. % towards this end.
% Several related works are working towards temporal consistency in video object-centric learning, all taking quite different approaches. 
For example, \citet{yu2024vonet} apply an object-wise sequential VAE to achieve consistency; \citet{Zhao2023ObjectCentricMO} and \citet{li2024reasoning} use an explicit memory buffer to maintain historical slot information and a transformer as a predictor using the memory buffer to predict the future; 
\citet{qian2023semantics} achieve temporal consistency by employing student-teacher distillation to establish semantic and instance correspondence over time; and
\citet{traub2023looping} use a recurrent network with a constancy prior \citep{gumbsch2021sparsely}.\looseness=-1
%Following semantic alignment, they utilize bipartite matching based on similarity measures to ensure instance-level consistency.
% \paragraph{Contrastive object-centric learning.} 

\section{Method}

\begin{figure*}[t]
    \centering
    \includegraphics[width=0.8\textwidth]{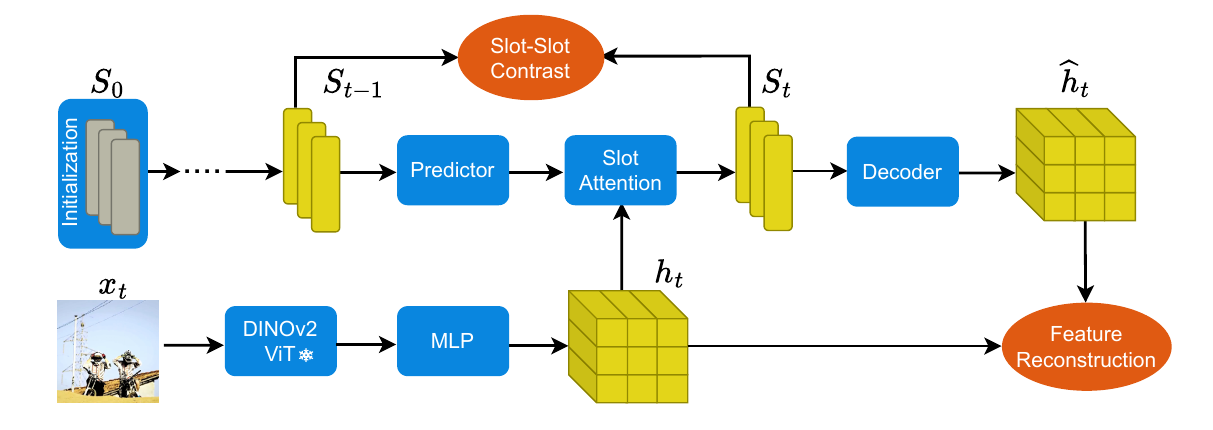}%
    \vspace{-1em}
    \caption{\method model architecture overview. For each frame, we extract patch features $h_t$ using DINOv2 ViT. These features are then used to update the previously initialized or predicted slots, resulting in new slots $S_t$. The model is trained by contrasting the current frame's slots $S_t$ with the slots from the previous frame $S_{t-1}$, and by reconstructing the patch features $h_t$.
    % \Andr{maybe it would be more interesting if we still show the difference between dense features by showing the grid and slots. Also, we need to improve names so that they are aligned with text, e.g. Slot-slot contrastive loss}
    }
    \label{fig:flow}
\end{figure*}
\label{sec:method}
% We propose \emph{slot-slot contrastive} loss function designed for making object-centric representations~(i.e., slots) consistent in time.
Our approach builds upon the existing input reconstruction-based video object-centric framework~\citep{kipf2022conditional, zadaianchuk2023objectcentric} by introducing a consistency loss that contrasts the slots across consecutive frames and thereby adapting the model to discover consistent representations. See \fig{fig:flow} for an overview of the \method architecture.\looseness=-1
% $s_{t-1}$ and $s_{t}$
% to be identified over time.
\subsection{Semantic Recurrent Slot Attention Module}
 Our model is an encoder-decoder object-centric architecture based on Slot Attention module (SA)~\cite{francesco2020object} with additional adaptations for sequential inputs similar to SAVi~\citep{kipf2022conditional}, while leveraging pre-trained semantic features as proposed by DINOSAUR~\citep{seitzer2023bridging}. The model consists of three main components: a pre-trained self-supervised dense feature encoder (e.g., DINOv2~\citep{oquab2023dinov2}), a Recurrent Slot Attention module that groups the encoder features into slots and models temporal slot updates, and a decoder that maps slots from each frame to reconstructions of the dense self-supervised features used as inputs. Next, we describe those components in more detail while explaining how to adapt them to the task of consistent object-centric representation learning.

%  To train the model we use feature reconstruction loss and temporal consistency loss functions.
% Object slots $ s_t = \{s^{1}_t, ...,s^{K}_t\}$ are extracted from the patch features $h_t$ of the self-supervised DINO model using a time-recurrent slot attention module~\citep{kipf2022conditional}, conditioned on the slots from the previous time step $t-1$. The model is trained to reconstruct the patch features $f_t$ corresponding to the current frame $x_t$. To ensure temporal consistency of the slot representations, we propose a novel contrastive slot-slot contrast loss that encourages the identity of the slots similarity across time.\looseness=-1
% \vspace{-0.5em}
Given a video frame $x_t, \ t \in \{1, 2, \dots, T\}$ and a pretrained, frozen self-supervised DINO model $f$ we first extract  $N$ patch features $g_t$,
\begin{equation}
g_t = f(x_t), \quad g_t \in \mathbb{R}^{N \times D}.
\end{equation}
As those frozen features are mostly semantic and are trained only on images, we further adapt them to the task of temporally consistent object discovery . Specifically, each feature vector $g_t$ is passed through a MLP $ g_{\psi}$,
\begin{equation}
  h_t = g_{\psi}(g_t),
\end{equation}
to adapt the frozen dense features for object-centric grouping (see ~\app{app:learned_features} for more details and visualizations). 
Based on the transformed encoder features $h_t$ and a set of slot representations of the previous timestep $S_{t-1}^p$, with $K$ slots $s_{t-1}^{k,p} \in S_{t-1}^p$, we use a recurrent grouping module to extract slot representations.
The Recurrent Slot Attention module comprises a grouping module $\text{C}_{\theta}$ and a
predictor module $\text{P}_{\omega}$. The former updates slot representations using the standard Slot Attention module~\citep{Locatello2020SlotAttention} on visual features $h_t$ from the encoder, while the latter captures temporal and spatial interactions between slots:
\begin{equation}
S_t^c = \text{C}_{\theta}(h_t, S_{t-1}^p), \quad S_t^p = \text{P}_{\omega}(S_t^c).
\end{equation}
Both slot-level representations, generated either by the grouping module $S_t^c$ or the predictor $S_t^p$, can be utilized for subsequent decoding or downstream task processing. In our implementation, the slot-level representations from the grouping module $S_t^c$  are employed for the decoding stage. From now on, we will refer to $S_t^c$ as $S_t$.

\begin{figure*}[t]
    \centering
    \includegraphics[width=0.9\textwidth]{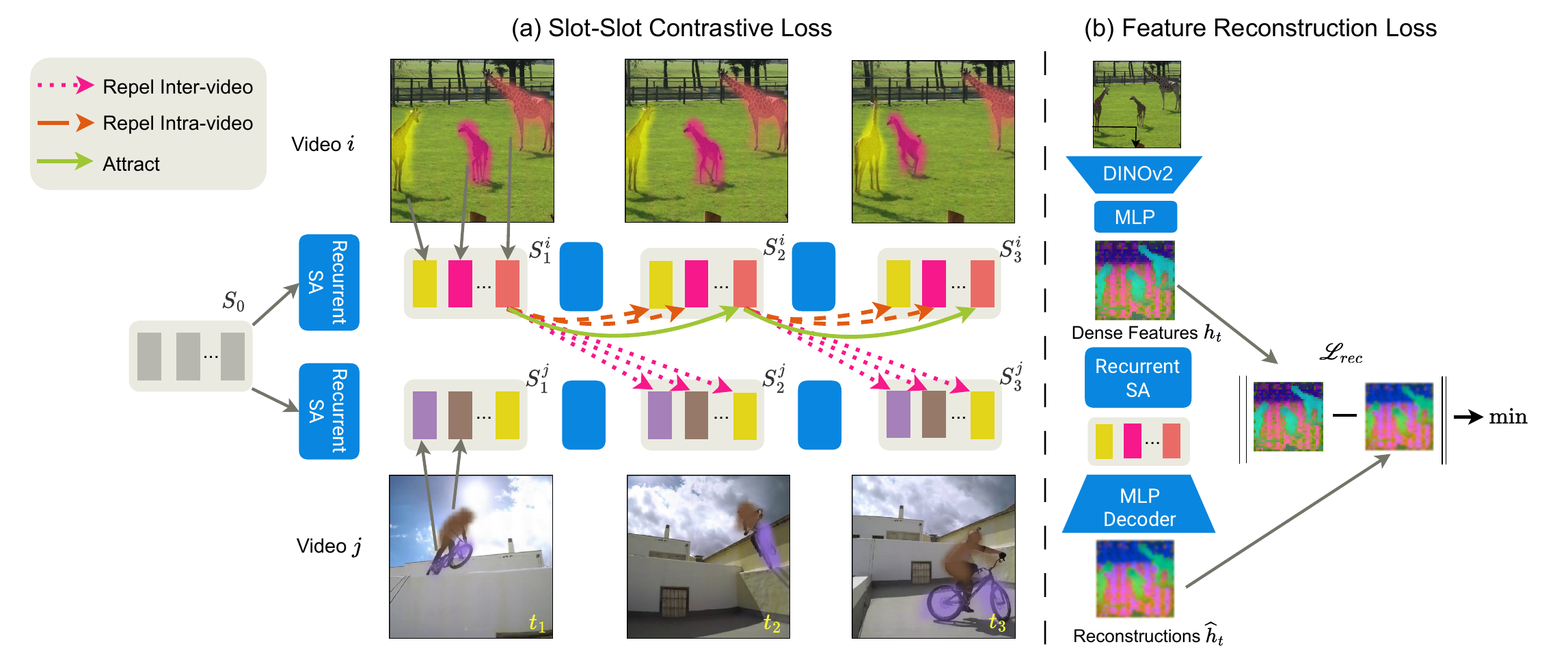}%
    \vspace{-.5em}
    \caption{Overview of the losses used in \method. (a) Our proposed temporal consistency objective, \emph{\loss}, operates on a batch of video sequences by enforcing temporal alignment across object slots. For each frame in the sequence, the model groups object features into specific slot representations ${S_{t}^i}$. The \loss then enforces temporal consistency by drawing the corresponding slot representations from adjacent frames closer, while simultaneously pushing apart all other slot representations in the batch---whether they come from different objects within the same video or from objects in other videos.
    (b) The feature reconstruction loss ensures informativeness of the learned slots by using them to reconstruct original DINOv2 features with an MLP decoder.
    }
    \label{fig:loss}
\end{figure*}

\paragraph{Temporal Slot Attention Initialization}
Importantly, we found that our setup benefits considerably from a learned initialization $S_0$, which can influence the efficiency of training across various objectives. Originally, \citet{Locatello2020SlotAttention} proposed a randomly sampled query initialization, where slots are sampled from the same Gaussian distribution with learned mean and variance. While such initialization allows different numbers of slots during inference, sampling from the same Gaussian distribution does not create a particularly favorable structure in slot-space. % to initializations that are close to each other. 
In this work, we use a straightforward learned initialization \citep{sajjadi2022object, jia2022improving} where a fixed set of initial slot vectors $S_0$ is learned for the entire dataset. Such initialization allows for learning dissimilar initialization queries that consistently attend to different objects.\looseness=-1
% There are two ways in which slots of the first frame $s_{0}$ can be initialized: randomly sampled queries initialization or learned queries initialization.  With random initialization, slots are sampled from the same Gaussian distribution with learned mean and variance. Meanwhile, with learned initialization, a fixed set of initial state vectors is learned for the entire dataset.
% These vectors are used to initialize the slots of the first frame for all videos.

Finally, for the reconstruction loss objective, we decode reconstructions $\hat{g}_t$ from all slots using the MLP decoder~\citep{seitzer2023bridging}.\looseness=-1

\subsection{Temporal Consistency through Slot Contrast}
Contrastive learning is flexible in supporting diverse data sources and loss function designs. By carefully defining positive and negative examples, we can craft robust loss objectives that effectively guide self-supervised representation learning~\citep{Chen2020SimCLR}. For instance, video contrastive methods like CVRL \citep{qian2021spatiotemporal}  leverage augmented video chunks to define positive (from the same video) and negative (from different videos) examples. In  object-centric learning \citet{didolkar2025ctrlo} employed a contrastive loss function to gain controllability over slot representations guided by language.
We propose a novel application of a contrastive loss for temporal consistency in object-centric slot representations.
In particular, we define positive samples as the representations of the same slot from two consecutive time steps within a video, while negative samples comprise all other slots across the batch between these time steps. An overview of the proposed loss is presented in~\fig{fig:loss}

\looseness=-1
% Our approach leverages the time dimension to define positive samples as the same slots from consecutive time steps within a video, while negative samples comprise all other slots across the batch between these time steps.\looseness=-1

\paragraph{Intra-Video Slot-Slot Contrastive Loss}
To force each slot to be consistent in time, we aim to learn slots that are similar in time while being maximally dissimilar to other slots.
Given the sets of slot representations $S_{t-1}$ and $S_{t}$ at time steps $ t-1 $ and $ t$, we want elements $s^i_{t-1} \in S_{t-1}$ to be close to the next-frame slots $s^i_{t}$ corresponding to the same object, while having maximal distance to the next-frame slots $s^k_{t}, k\neq i$ corresponding to other objects in the video. The corresponding InfoNCE contrastive loss~\citep{oord2018representation} is defined as $
\mathcal{L}_{\text{intra}} = \frac{1}{K} \sum_{i=1}^{K} \ell^{\text{\,intra}}_{i}
$ with
\begin{equation}
\small
\ell^{\text{\,intra}}_{i} = -\log \frac{\exp(\text{sim}(s^{i}_{t-1}, s^{i}_{t}) / \tau)}{\sum_{k=1}^K \mathbbm{1}_{[k \neq i]} \exp(\text{sim}(s^{i}_{t-1}, s^{k}_{t}) / \tau)},
\end{equation}
where $K = |S_{t}|$ is a number of slots per frame, \(\text{sim}(u, v) = \frac{u^\top v}{\|u\|_2 \|v\|_2}\) is the
cosine similarity, \(\mathbbm{1}_{[.]} \) is an indicator excluding the self-similarity of the slot \( s_i \) from the denominator, and \( \tau > 0 \) is a temperature parameter.
% \begin{equation}
%     \mathcal{L}_{\text{intra}} = -\log\frac{\exp(s^i_{t} \cdot s^i_{t+1}/\tau)}{\sum_{j=1}^K \exp(s^i_{t} \cdot s^{j}_{t+1}/\tau)},
% \end{equation} where $K = |S_{t+1}|$ is a number of slots in one video and $\tau$ is a temperature hyperparameter.

While being a desirable property, intra-video slot contrast can be achieved simply by amplifying the differences between slots in the SA module's first frame initialization $S_0$.
% simply by ignoring the original video and making slots contrastive by emphasizing the original difference in the initialization of the SA module.
To encourage a stronger focus on video content and instance specificity of the representations, we propose a further improvement over this loss by extending the negative contrast set.\looseness=-1
%Next, we propose further improvement over such loss by extending the negative contrast set and thus encouraging greater separation between slot representations while maintaining their informativeness.\looseness=-1
% to include slots from all frames within a batch, encouraging greater separation between slot representations.

% \paragraph{Intra-inter video slot-slot contrastive loss}
\paragraph{Batch Video Slot-Slot Contrastive Loss}
To leverage the benefits of larger contrast sets and prevent degenerate solutions relying solely on the initialization of slots, we exploit the fact that the whole batch of videos can be considered a large set of primarily unique object representations.
%Thus, to learn temporally consistent object representations, we can enhance the representations' contrast both within a video (between object representations) and across different videos in the same batch. For this, we extend the set of negatives to include slots not only from the current frame at time $t$ and the subsequent frame at time $t+1$, but also from all frames within the batch of videos being processed together.
Consequently, we enhance contrast within a video and between videos by including negative slots from the current and subsequent frames of all videos in the batch.
Correspondingly, we define our slot-slot contrastive loss as $
\mathcal{L}_{\text{ssc}} = \frac{1}{B\cdot K} \sum_{j=1}^{B} \sum_{i=1}^{K} \ell_{i, j}^{\text{\,ssc}}
$ and
\begin{equation}
\small
    \ell_{i, j}^{\text{\,ssc}} = -\log\frac{\exp(\text{sim}(s^{i, j}_{t-1}, s^{i, j}_{t})/\tau)}{ \sum\limits_{b=1}^{B}  \sum\limits_{k=1}^{K}  \mathbbm{1}_{[k,b \neq i,j]} \exp(\text{sim}(s^{i, j}_{t-1}, s^{k,b}_{t})/\tau)},
\end{equation} where $B$ is a number of videos in the batch  and $ s^{i,j}_t $ denotes the $ i $-th slot of the $ j $-th video at time $ t $. For more details on slot-slot contrastive loss implementation, see \app{app:loss_details}.\looseness=-1

%This modification enables a comprehensive comparison of all slots within the batch by incorporating slots from different batch videos as additional negatives in the contrastive loss.
We find that this approach significantly enhances the effectiveness of the slot-slot contrastive loss. Furthermore, since all videos in the batch are processed with the same initial state $S_0$, this loss function avoids suboptimal solutions that rely solely on the uniqueness of the initialization, instead encouraging object discovery as the basis for contrast.

% Motivated by the observation that contrastive losses perform better with larger contrast sets, and to avoid degenerate solutions when the contrast is achieved only by relaying on original slots initialization, we modify our contrastive loss to achieve both constrastiveness of each object representation within video as well as contrastiveness of the object representations though different videos.  the similarity matrix $\mathbf{A}$ to include not only the slots for the current frame at time step $t$ and the subsequent frame at time step $t+1$, but also the slots from all frames within the batch of videos that are processed together.

\paragraph{Final Loss}  To encourage scene decomposition we use a feature reconstruction loss, similar to  DINOSAUR~\cite{seitzer2023bridging} and VideoSAUR~\cite{zadaianchuk2023objectcentric}. Our final loss function combines the reconstruction loss with our proposed contrastive loss $\mathcal{L}_{\text{ssc}}$, weighted by the hyperparameter $\alpha$ (see \tab{tab:implementation_details} for details on how the hyperparameters are set):
\begin{equation}
\small
    \mathcal{L} = \sum_{t=1}^{T-1} \mathcal{L}_{\text{rec}}(\mathrm{h_t}, \mathrm{\hat{h}_t}) + \alpha \mathcal{L}_{\text{ssc}}(S_{t-1}, S_{t}).
\end{equation}
\looseness=-1

\section{Experiments}
\label{sec:experiments}

\begin{table}
\centering
    \caption{Consistent object-discovery performance of \method in comparison with SAVi, STEVE, VideoSAUR on MOVi-C, MOVi-E, and YouTube-VIS datasets. VideoSAURv2 is an improved version of the VideoSAUR trained on DINOv2 features.  Both metrics are computed for the whole video (24 frames for MOVi, up to 76 frames for YouTube-VIS).}
    \label{tab:movi_youtube_metrics}
    \vspace{-.5em}
    \normalsize% Change to smaller font size
    \setlength{\tabcolsep}{3pt} % Reduce the space between columns
    \adjustbox{max width=\linewidth}{
   \begin{tabular}{l@{\,\,}cccccc} % Adjust column layout
    \toprule
       & \multicolumn{2}{c}{\textbf{MOVi-C}} & \multicolumn{2}{c}{\textbf{MOVi-E}} & \multicolumn{2}{c@{}}{\textbf{YouTube-VIS}} \\
    \cmidrule(lr){2-3} \cmidrule(lr){4-5} \cmidrule(lr){6-7}
    & \!\!\textbf{FG-ARI} $\uparrow$ & \textbf{mBO} $\uparrow$ & \textbf{FG-ARI} $\uparrow$ & \textbf{mBO} $\uparrow$ & \textbf{FG-ARI} $\uparrow$ & \textbf{mBO} $\uparrow$ \\
    \midrule
    % Feat. Rec.  & 49.7 & 27.4 & 79.8 & 28.4 & 35.2 & 31.4 \\
    % SOLV & - & - & 80.8 & - & - & - \\
    SAVi~\citep{kipf2022conditional}   & 22.2 & 13.6 & 42.8 & 16.0 & - & - \\
    STEVE~\citep{Singh2022STEVE}   & 36.1 & 26.5 & 50.6 & 26.6 & 15 & 19.1 \\
    VideoSAUR~\citep{zadaianchuk2023objectcentric}   & 64.8 & \textbf{38.9} & 73.9 & \textbf{35.6} & 28.9 & 26.3\\
    VideoSAURv2   & - & - &77.1 & 34.4 & 31.2 & 29.7\\
    \rowcolor{TableColor} \method & \textbf{69.3 }& 32.7 & \textbf{82.9} & 29.2 & \textbf{38.0} & \textbf{33.7} \\
    \bottomrule
\end{tabular}}
\vspace{-.5em}
\end{table}

We evaluate our method's temporal consistency on two downstream tasks: object discovery and latent object dynamics prediction. Our experiments address three main questions:
(1)~How does our model compare to state-of-the-art methods in both temporal consistency and scene decomposition?
(2)~How effective are our model's learned representations for the challenging downstream task of object dynamics prediction and for object tracking under full occlusions?
(3)~How important are the different components of our model and loss function for temporal consistency?\looseness=-1

% We conducted experiments to evaluate our method's effectiveness in maintaining temporal consistency on two downstream tasks: (1) object discovery and (2) video prediction.

\paragraph{Datasets} To evaluate our method in the controlled setting, we use  MOVi-C and MOVi-E synthetic datasets generated by Kubric~\citep{Greff2021Kubric}. MOVi-C includes richly textured everyday objects, featuring up to 11 objects per scene, while MOVi-E expands this to 23 objects and introduces basic linear camera motion. In addition, to study the scalability of our method to real-world data,  we evaluate our method on the real-world YouTube-VIS 2021 (YTVIS21) video dataset~\citep{vis2021}.  YTVIS21 is an unconstrained, real-world dataset sourced from YouTube, capturing a diverse range of scenes~(for more details, see \app{app:dataset_details}).\looseness=-1

\paragraph{Metrics}
Similar to other object-centric video methods~\citep{elsayed2022savi++, kipf2022conditional, Singh2022STEVE, zadaianchuk2023objectcentric}, to evaluate consistent object discovery, we use the video foreground adjusted rand index (FG-ARI)~\cite{greff2019multi}, measuring how well objects are split. In addition, we evaluate the sharpness of masks using the video intersection over union with mean best overlap matching (mBO) metric~\citep{seitzer2023bridging,zadaianchuk2023objectcentric}. Both metrics are computed \emph{over the full video} and thus reflect how consistent object discovery is. In addition, to investigate the effects of the temporal consistency inductive bias on the per-frame object discovery itself, we use per-frame FG-ARI~(image FG-ARI), which we independently compute for each frame and average afterwards. More details can be found in \app{app:metrics_details}.

Finally, when evaluating how well object-centric representations perform for object dynamics prediction (see \cref{subsec:dynamics-prediction}), we employ the same evaluation metrics as in the object discovery task: FG-ARI and mBO. This time, however, these metrics are computed by comparing the predicted masks (obtained by decoding the predicted slots~\citep{wu2023slotformer}) with the ground-truth future masks.%\looseness=-1

\subsection{Object Discovery}

\paragraph{Implementation Details}
We employ the DINOv2 model as our feature encoder, using ViT-S/14 for the MOVi-C dataset and ViT-B/14 for MOVi-E and YTVIS21. The slot dimension is set to 128 for MOVi-E and 64 for both MOVi-C and YTVIS21. For the MOVi datasets, we use a resolution of $(336, 336)$, generating $24\!\times\!24$ patches yielding $576$ ViT tokens, while for YTVIS21, a resolution of $(518, 518)$ yields $1369$ tokens. Full details are provided in \app{app:training_details}.\looseness=-1

\paragraph{Baselines} We compare \method against the previously proposed SAVi~\citep{kipf2022conditional} and STEVE~\citep{Singh2022STEVE} that employ an image reconstruction objective and with the state-of-the-art method VideoSAUR~\citep{zadaianchuk2023objectcentric} that uses self-supervised feature reconstructions. Additionally, for a fair comparison, we trained a modification of VideoSAUR with DINOv2 features (referred to as VideoSAURv2). The implementation details are provided in~\app{app:baseline_details}. In addition, to assess how closely \method approaches supervised methods, we compared it with SAM2~\citep{ravi2024sam2} as a supervised zero-shot baseline for temporal consistency and to weakly-supervised by depth SAVi++~\citep{elsayed2022savi++} method. The results are in \app{app:baseline_details}. 
% We also compared our method to SAVi++~\citep{elsayed2022savi++} weakly-supervised method (see \app{app:baseline_details}).

%\paragraph{Baselines} We compare \method against the previously proposed SAVi~\citep{kipf2022conditional} and STEVE~\citep{Singh2022STEVE} that employ an image reconstruction objective and with the state-of-the-art method VideoSAUR~\citep{zadaianchuk2023objectcentric} that uses self-supervised feature reconstructions. Additionally, for a fair comparison, we trained a modification of VideoSAUR with DINOv2 features (referred to as VideoSAURv2) using the same video resolution as our method.
%Interestingly, the default VideoSAURv2 configuration was not preforming well on the simplest MOVi-C dataset (while other dataset were working better than VideoSAUR without additional changes), thus we additionally explored possible parameters choice in~\App{app:baseline_details}.
%Interestingly, default VideoSAURv2 configuration was not preforming well on the simplest MOVi-C dataset, thus we additionally explored possible parameters there in~\App{}.

\begin{figure*}
    \centering
    \includegraphics[width=\textwidth]{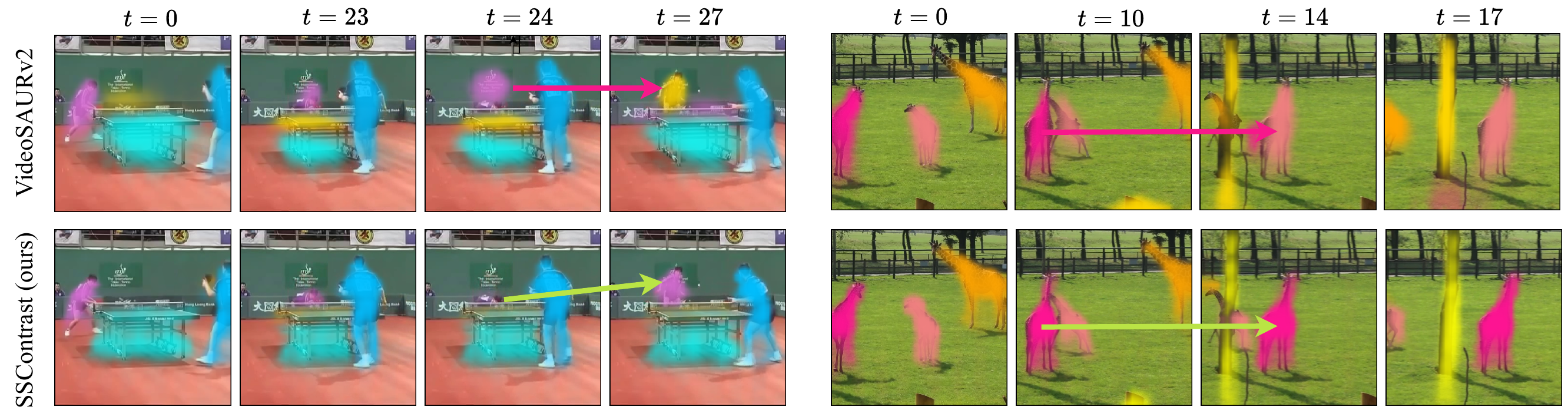}\vspace{-.5em}
    \caption{Qualitative comparison with VideoSAURv2 on YouTube-VIS dataset. In challenging situations~(e.g., almost full occlusions at $t=24$ of the 1st video and $t=14$ of the 2nd video), VideoSAURv2 reassigns slots to different objects (pink arrows), whereas \method consistently assigns slots to the same object (green arrows). Note that the colors of the masks are matched manually for better visual comparison.}
    \label{fig:comparison_with_videosaur}
\end{figure*}

\paragraph{Temporally Consistent Object Discovery (\Tab{tab:movi_youtube_metrics} \& \Fig{fig:comparison_with_videosaur})}
% The results are presented in~\tab{tab:movi_youtube_metrics}.
\method significantly outperforms both SAVi and STEVE by a wide margin. When compared to VideoSAUR using its default parameters, our approach demonstrates higher consistency in terms of video FG-ARI scores. Compared to VideoSAUR and VideoSAURv2, \method achieves superior video scene decomposition (measured by FG-ARI scores). However, on synthetic datasets~\method's masks are less sharp~(as reflected by mBO). Notably, on the most challenging real-world YouTube-VIS data, our method surpasses both versions of VideoSAUR, achieving better performance on FG-ARI ($+6.8$) and mBO ($+4$). This shows that, given a large enough resolution and natural data inputs well aligned with DINOv2, \method can decompose unconstrained videos into consistent object representations. More examples are illustrated in \app{app:additional_examples}.\looseness=-1
% However, for the MOVi-C and MOVi-E datasets, we observe lower mBO scores, likely due to VideoSAUR utilizing more image tokens.

\begin{table}[t]
\centering
\caption{Quantitative Results on MOVi-E in terms of per-frame Image FG-ARI. The methods are grouped by the target data they train on: only images ($\mathcal{I}$), videos with motion segmentation annotations ($\mathcal{V} + \mathcal{M}$), and only videos ($\mathcal{V}$). }
\small %
\setlength{\tabcolsep}{6pt} %
\vspace{-.5em}
\begin{tabular}{l l l c} %
    \toprule
    \multirow{2}{*}{} & \textbf{Model} & \textbf{Objective} & \textbf{Image} \\ &&&\textbf{FG-ARI} $\uparrow$ \\
    \midrule
    \multirow{3}{*}{\rotatebox{90}{$\mathcal{I}$}} & LSD~\citep{jiang2023object} & Image Rec. & 53.4 \\
    & SlotDiffusion~\citep{wu2023slotdiffusion} & Image Rec. & 60.0 \\
    & DINOSAUR~\citep{seitzer2023bridging} & Image Rec. & 65.1 \\
    \midrule
    \multirow{3}{*}{\rotatebox{90}{$\mathcal{V}+\mathcal{M}$}} & MoToK~\citep{bao2023object} & +Mot. Seg. & 66.7 \\
    & Safadoust et al.~\citep{safadoust2023multi} & +GT Flow & 78.3 \\
    & DIOD~\citep{Kara_2024_CVPR} & +Mot. Seg. & 82.2 \\
    \midrule
    \multirow{4}{*}{\rotatebox{90}{$\mathcal{V}$}} & STEVE~\citep{Singh2022STEVE} & Video Rec. & 54.1 \\
    & VideoSAUR~\citep{zadaianchuk2023objectcentric} & Temp. Sim. & 78.4 \\
    & SOLV~\citep{aydemir2023self} & Mid. Fr. Pred. & 80.8 \\
       & \cellcolor{TableColor}\method & \cellcolor{TableColor}Slot Contrast & \cellcolor{TableColor} \textbf{84.8} \\
    \bottomrule
\end{tabular}
\label{tab:movi_e_image}
\end{table}

% To ensure a fair comparison, we also retrained VideoSAUR with DINOv2 features (referred to as VideoSAURv2).
\paragraph{Per-Frame Scene Decomposition~(\Tab{tab:movi_e_image})  }

Previous research has shown the effectiveness of specific inductive biases and training objectives for unsupervised object discovery, such as reconstructing in semantic space or leveraging motion cues in video data.
Building on this, we demonstrate that the contrastive nature of our temporal consistency loss function yields improved scene decomposition as a byproduct.
This occurs because our loss function encourages the model to learn consistent feature representations for objects across frames, leading to an adaptive process where dense features become more contrastive, thereby enhancing object discovery in individual frames.

We compare our method with prior approaches in terms of per-frame object discovery, using the image FG-ARI metric for evaluation. Specifically, we compare three categories of methods: image-based, video-based, and methods that use videos with additional motion cues. Image-based methods use only images as a target (feature reconstruction based DINOSAUR~\citep{seitzer2023bridging} and diffusion-based LSD~\citep{jiang2023object} and SlotDiffusion~\citep{wu2023slotdiffusion} methods). Video-based methods use only videos as targets: STEVE~\citep{Singh2022STEVE} reconstructs current frame features, SOLV~\citep{aydemir2023self} predicts middle frame features, and VideoSAUR~\citep{zadaianchuk2023objectcentric} predicts temporal feature-similarities. Finally, we also compare with weakly-supervised methods using motion masks~\citep{bao2023object, Kara_2024_CVPR} or ground truth (GT) optical flow~\citep{safadoust2023multi}.

The results on MOVi-E dataset are presented in~\tab{tab:movi_e_image}, with comparisons across additional datasets provided in \app{app:image}. Using temporal signals from the video using feature reconstruction is better than object discovery based on images. Next, additional objectives that exploit the temporal structure of the videos allow even better scene decomposition. Notably, our method, which combines a feature reconstruction objective with a simple contrastive objective, leads to state-of-the-art performance reaching $84.8$ per-frame FG-ARI, outperforming methods~\citep{bao2023object, Kara_2024_CVPR} that use motion segmentation masks for object discovery.\looseness=-1
% \vspace{-1em}
\paragraph{Robustness to Full Occlusions}
To evaluate our method's robustness in handling complete object occlusions---a challenging scenario for maintaining consistency---we conduct experiments using a targeted subset of the MOVi-C dataset that contains sequences where objects are fully occluded.
% To assess our method's robustness in handling complete object occlusions—a particularly challenging scenario for maintaining consistency—we conduct experiments on a targeted subset of the MOVi-C dataset that focuses exclusively on fully occluded object sequences.
% The MOVi-C dataset provides visibility scores for each object in each frame, indicating the number of pixels the object occupies. Using these scores, we refine the validation set to include only sequences meeting the following conditions: an object initially appears with a visibility score of at least $\mathrm{n}$ pixels, then becomes fully occluded (visibility score drops to 0 pixels), and subsequently reappears with a visibility score of at least $\mathrm{n}$ pixels.
% % More details and the visualization of the proposed subset can be found in \app{app:occusions}.
% To avoid including very small objects or visual artifacts, we set $\mathrm{n}$ to a minimum of $400$ pixels (less than 1\% of the image pixels).
% % After applying this filtering criterion, we obtain a dataset of 60 sequences where objects undergo complete occlusion and reappearance.
For evaluation, we retain only the ground-truth masks for the objects that experience occlusion.\looseness=-1

\begin{figure}[t]
    \centering
    \includegraphics[width=\columnwidth]{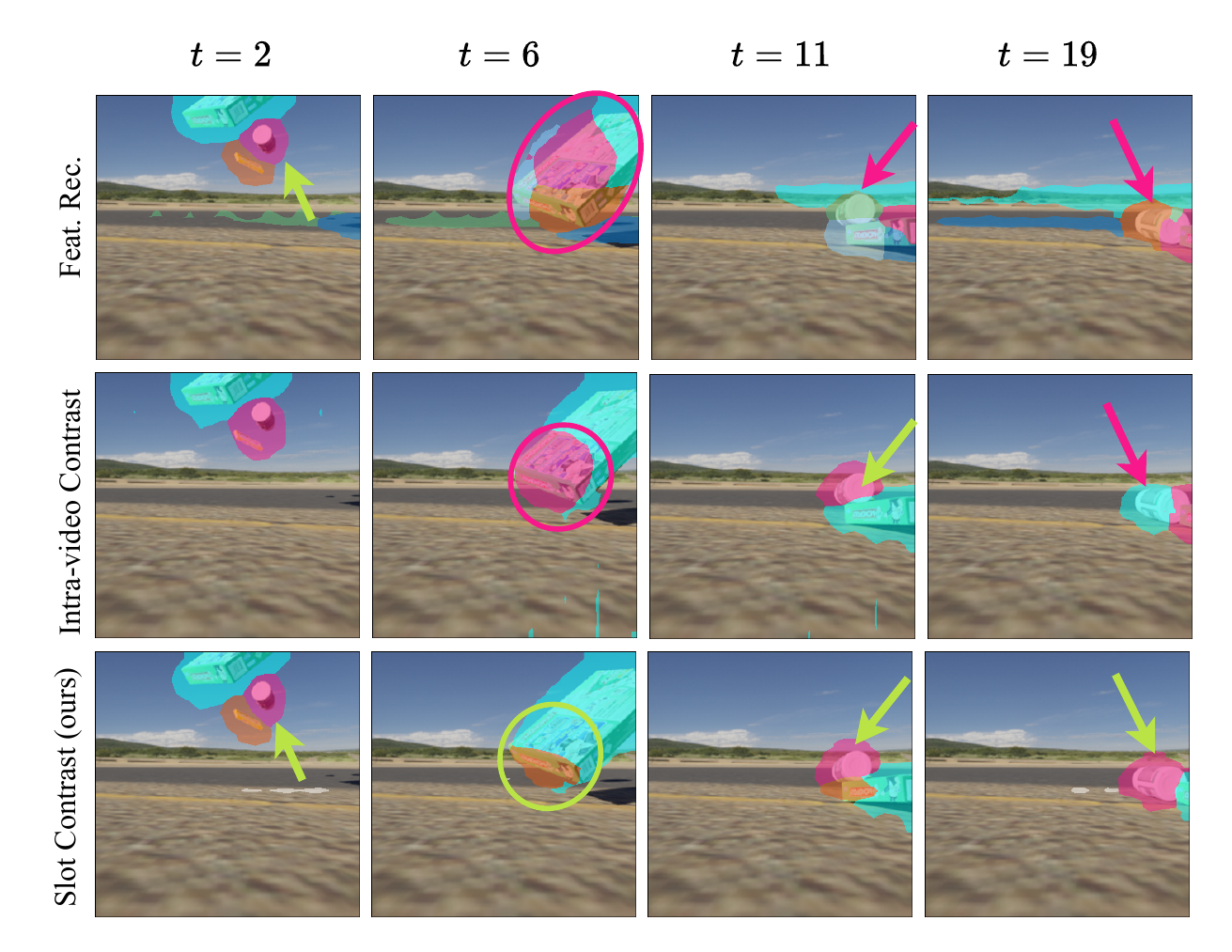}% 
    \caption{
Comparison of the Feature Reconstruction (Feat. Rec.) baseline, the slot-slot contrastive loss using only slots from the same video as the contrast set (Intra-video Contrast), and \method on the MOVi-C dataset. 
    \looseness=-1}
    \label{fig:comparison_movic}
\end{figure}

We find that the feature reconstruction baseline achieves only $16\%$ mBO vs.\ our method obtains $21\%$ mBO on fully occluded objects.
Our results suggest that \method significantly enhances consistency during object disappearances and reappearances.
%, improving from a feature reconstruction baseline~($16\%$ vs. $21\%$ mBO on fully occluded objects) and demonstrating object permanence capabilities.
We refer the reader to \fig{fig:comparison_movic} and \app{app:movi-c-occluded visualizations} for visual examples and more details.
% This demonstrates emerging object permanence capabilities, outperforming the baseline model.

% \begin{table}[h]
% \centering
% \caption{Quantitative Results on MOVi-E in terms of per-frame Image FG-ARI.}
% \small % Increase font size
% \setlength{\tabcolsep}{6pt} % Adjust space between columns for readability
% \renewcommand{\arraystretch}{1.2} % Increase vertical space in rows
% \begin{tabular}{llc} % Adjust column layout
%     \toprule
%     & \textbf{Objective} & \textbf{FG-ARI} $\uparrow$ \\
%     \midrule
%     LSD~\citep{jiang2023object} & Image Rec. &53.4 \\
%     SlotDiffusion~\citep{wu2023slotdiffusion} & Image Rec. &60.0 \\
%     DINOSAUR~\citep{seitzer2023bridging} &Image Rec.& 65.1 \\

%     %vDINOSAUR &Video Rec. & 76.3 \\
%     \midrule
%     MoToK~\citep{bao2023object} & +Mot. Seg. & 66.7 \\
%     Safadoust et al.~\citep{safadoust2023multi} & +GT Flow & 78.3 \\
%     DIOD~\citep{Kara_2024_CVPR} & +Mot. Seg. & 82.2 \\
%     \midrule
%         STEVE~\citep{Singh2022STEVE} & Video Rec.& 54.1 \\VideoSAUR~\citep{zadaianchuk2023objectcentric} & Temp. Sim. & 78.4 \\
%     SOLV~\citep{aydemir2023self} & Mid. Fr. Pred. & 80.8 \\
%     \rowcolor{TableColor} \method & Slot Contrast & \textbf{84.8} \\

%     \bottomrule
% \end{tabular}
% \label{tab:movi_e_image}
% \end{table}

\subsection{Object Dynamics Prediction}
\label{subsec:dynamics-prediction}
\paragraph{Setup} To evaluate performance on the task of predicting object dynamics, we train a dynamics module using the object-centric representations inferred by a pretrained object-centric model. For this dynamics module, we select SlotFormer~\cite{wu2023slotformer}, which predicts the slots autoregressively for $K$ rollout steps based on the slots inferred from $T$ burn-in frames preceding the prediction horizon. In our setup, we use 14 burn-in frames and 10 rollout steps. Since SlotFormer is trained independently from the object-centric model, we first train the latter, subsequently extending the datasets with the inferred slots for each frame. This approach avoids the computational complexity of training SlotFormer by removing the necessity to encode frames into the slot space at each training step. A brief introduction to SlotFormer and the implementation details can be found in \app{app:slotformer}. 

\begin{figure}[b]
    \centering
    \vspace{-1em}
    \includegraphics[width=.99\columnwidth]{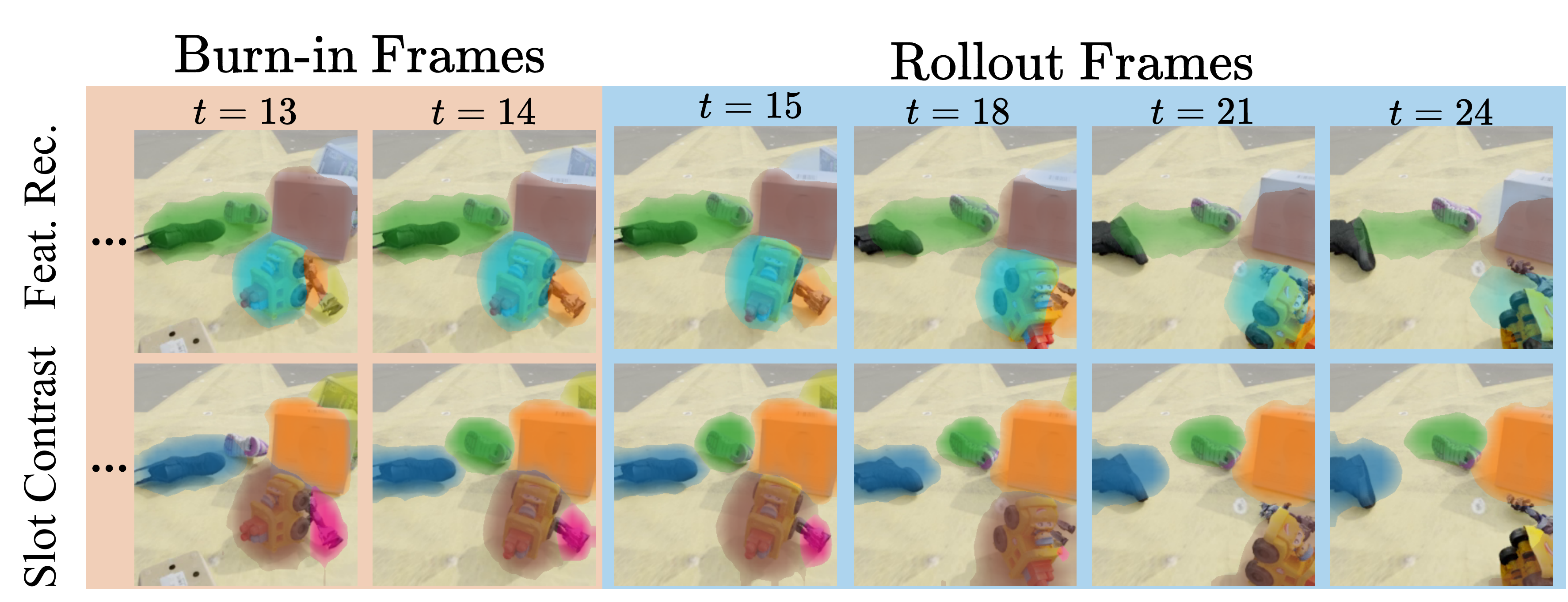}%
    \vspace{-.5em}
    \caption{Object dynamics prediction task on MOVi-C using \method slots using SlotFormer~\citep{wu2023slotformer}.}
    \label{fig:slotformer_figure}
\end{figure}
\paragraph{Baselines} We compare SlotFormer~\citep{wu2023slotformer} trained on object-centric representations derived from a model trained using only feature reconstruction loss with SlotFormer trained on representations from \method. Both models perform reconstruction in feature space rather than pixel space, so we use only the slot reconstruction loss for training.\looseness=-1
\begin{table}[tb]
\centering
    \caption{Downstream task of predicting object dynamics. Comparison of predictions made by SlotFormer based on representations obtained from \method and from Feature Reconstruction.}
    \vspace{-0.5em}
    \setlength{\tabcolsep}{2pt} % Reduce the space between columns
    \adjustbox{max width=\linewidth}{
   \begin{tabular}{l@{}cccccc} % Adjust column layout
    \toprule
       & \multicolumn{2}{c}{\textbf{MOVi-C}} & \multicolumn{2}{c}{\textbf{MOVi-E}} & \multicolumn{2}{c@{}}{\textbf{YouTube-VIS}} \\
    \cmidrule(lr){2-3} \cmidrule(lr){4-5} \cmidrule(lr){6-7}
    & \textbf{FG-ARI} $\uparrow$ & \textbf{mBO} $\uparrow$ & \textbf{FG-ARI} $\uparrow$ & \textbf{mBO} $\uparrow$ & \textbf{FG-ARI} $\uparrow$ & \textbf{mBO} $\uparrow$ \\
    \midrule
    Feat. Rec. + SF & 50.7 & 25.9 & 70.6 & 24.3 & 27.4 & 28.9 \\
    \rowcolor{TableColor} Ours + SF & 63.8 & 26.1 & 70.5 & 24.9 & 29.2 & 29.6 \\
    \bottomrule
\end{tabular}
}
    \label{tab:odp_metrics}
%\vspace{-2em}
\end{table}

\paragraph{Quality of Predicted Masks (\Tab{tab:odp_metrics} and \Fig{fig:slotformer_figure})}
% The results of the experiment can be found in \tab{tab:odp_metrics}.
We note that our model has a significantly better FG-ARI on MOVi-C, while mBO is comparable to that of the baseline. On MOVi-E, the performance of our model is comparable to that of the baseline, which highlights the difficulty of adapting to videos with camera motion. There is also a slight improvement in FG-ARI on YTVIS21, while mBO remains comparable. It is worth noting that predicting the motion in this dataset is especially challenging, given the large diversity of possible scenarios.
% This dataset also includes a notable number of entities that do not follow simple physical rules but instead behave as actors.

\subsection{Analysis}
In this section, we investigate key components of our approach, including the impact of the contrastive loss and the type of slot initialization. In addition, we study how effective \method is in automatically shutting down slots in correspondence to the scene's complexity.
\label{sec:analysis}
%in progress
\paragraph{Ablation of Loss Components~(\Tab{tab:batch_contrast} and   \Fig{fig:comparison_movic})}
To demonstrate the value of the proposed slot-slot contrastive loss, we carry out an ablation study, comparing it with the feature reconstruction loss~\citep{seitzer2023bridging} and the intra-video contrastive loss, which contrasts slot representations in a single video.
% We explored the impact of having all batch elements within the same contrast set.
% The results are summarized in \tab{tab:batch_contrast}.
Using the intra-video contrastive loss yields improvements over the feature reconstruction baseline~($+5.1$ FG-ARI and $+1.5$ mBO on MOVi-C). However, we observe that in more challenging situations, the intra-video contrastive loss leads to failure cases such as shutting down too many slots~(see \fig{fig:comparison_movic}).  Next, we observe that by extending the contrast to the full batch of videos, \method learns more consistent representations ($+19.6$ FG-ARI and $+5.3$ mBO). %This approach makes the learning task more challenging, which helps prevent the model from relying on superficial patterns, such as slot initializations or object positions.
This change increases the difficulty of the learning task, which prevents the model from relying on superficial patterns like slot initializations or object positions.
% In \fig{fig:comparison_movic}, we
% The quantitative results are summarized in \tab{tab:batch_contrast}.

\begin{table*}[ht]

\centering
\caption{Ablation of loss components used by \method on MOVi-C, MOVi-E, and YouTube-VIS Datasets.}
\vspace{-0.5em}
\small % Reduce font size
\setlength{\tabcolsep}{3pt} % Reduce space between columns
% \renewcommand{\arraystretch}{0.8} % Reduce vertical space in rows: BUT it also makes the background filling to be chopped of at the top.
% \begin{tabular}{@{}lcccccc} % Adjust column layout
%     \toprule
%     & \multicolumn{2}{c}{\textbf{MOVi-C}} & \multicolumn{2}{c}{\textbf{MOVi-E}} & \multicolumn{2}{c}{\textbf{YouTube-VIS}} \\
%     \cmidrule(lr){2-3} \cmidrule(lr){4-5} \cmidrule(lr){6-7}
%     & \textbf{FG-ARI} $\uparrow$ & \textbf{mBO} $\uparrow$ & \textbf{FG-ARI} $\uparrow$ & \textbf{mBO} $\uparrow$ & \textbf{FG-ARI} $\uparrow$ & \textbf{mBO} $\uparrow$ \\
%     \midrule
%     Feat. Rec. (no contrast)  & 49.7 & 27.4 & 79.8 & 28.4 & 35.3 & 31.4 \\
%         \midrule
%     \method ($\mathcal{L}_{\text{intra}}$ contrast) & 54.8  & 28.9 & 78.7 & 29.1 & 35.7 & 33.6 \\
%     \rowcolor{TableColor} \method & 69.3 & 32.7  &82.9 & 29.2& 38.0 & 33.7 \\
%     \bottomrule
% \end{tabular}
\begin{tabular}{cccccccccc}
    \toprule
    Feat. Rec. & Intra-video Contrast & Slot-Slot Contrast && \multicolumn{2}{c}{\textbf{MOVi-C}} & \multicolumn{2}{c}{\textbf{MOVi-E}} & \multicolumn{2}{c}{\textbf{YouTube-VIS}} \\
    \cmidrule(lr){5-6} \cmidrule(lr){7-8} \cmidrule(lr){9-10}
    $\mathcal{L}_{\text{rec}}$ & $\mathcal{L}_{\text{intra}}$ & $\mathcal{L}_{\text{ssc}}$ &  & \textbf{FG-ARI} $\uparrow$ & \textbf{mBO} $\uparrow$ & \textbf{FG-ARI} $\uparrow$ & \textbf{mBO} $\uparrow$ & \textbf{FG-ARI} $\uparrow$ & \textbf{mBO} $\uparrow$ \\
    \midrule
    \checkmark & & & & 49.7 & 27.4 & 79.8 & 28.4 & 35.3 & 31.4 \\
    % \midrule
    \checkmark & \checkmark & & & 54.8  & 28.9 & 78.7 & 29.1 & 35.7 & 33.6 \\
    \rowcolor{TableColor} \checkmark &  & \checkmark & & 69.3 & 32.7  & 82.9 & 29.2 & 38.0 & 33.7 \\
    \bottomrule
\end{tabular}

% \begin{tabular}{ccccccccc@{}}
%     \toprule
%      $\mathcal{L}_{\text{rec}}$ & $\mathcal{L}_{\text{intra}}$ & $\mathcal{L}_{\text{intra-inter}}$  \multicolumn{2}{c}{\textbf{MOVi-C}} & \multicolumn{2}{c}{\textbf{MOVi-E}} & \multicolumn{2}{c}{\textbf{YouTube-VIS}} \\
%     \cmidrule(lr){6-7} \cmidrule(lr){9-7} \cmidrule(lr){10-11}
%     & & & & \textbf{FG-ARI} $\uparrow$ & \textbf{mBO} $\uparrow$ & \textbf{FG-ARI} $\uparrow$ & \textbf{mBO} $\uparrow$ & \textbf{FG-ARI} $\uparrow$ & \textbf{mBO} $\uparrow$ \\
%     \midrule
%     \checkmark & & & 49.7 & 27.4 & 79.8 & 28.4 & 35.3 & 31.4 \\
%     \midrule
%     \checkmark & \checkmark &  & 54.8  & 28.9 & 78.7 & 29.1 & 35.7 & 33.6 \\
%     \rowcolor{TableColor} \checkmark & \checkmark & \checkmark &  69.3 & 32.7  & 82.9 & 29.2 & 38.0 & 33.7 \\
%     \bottomrule
% \end{tabular}

\label{tab:batch_contrast}
\vspace{-1em}
\end{table*}

\begin{table}
\centering
\caption{Comparison of the random initialization (RI) and learned initialization (LI) techniques.}
\vspace{-0.5em}
\setlength{\tabcolsep}{3pt} % Reduce space between columns
\adjustbox{max width=\linewidth}{
\begin{tabular}{l@{\,}cccccc} % Adjust column layout
    \toprule
    & \multicolumn{2}{c}{\textbf{MOVi-C}} & \multicolumn{2}{c}{\textbf{MOVi-E}} & \multicolumn{2}{c}{\textbf{YouTube-VIS}} \\
    \cmidrule(lr){2-3} \cmidrule(lr){4-5} \cmidrule(lr){6-7}
    & \textbf{FG-ARI} $\uparrow$ & \textbf{mBO} $\uparrow$ & \textbf{FG-ARI} $\uparrow$ & \textbf{mBO} $\uparrow$ & \textbf{FG-ARI} $\uparrow$ & \textbf{mBO} $\uparrow$ \\
    \midrule
    Feat.~Rec.~(RI) & 45.3  & 27.2 & 71.1 & 28.3 & 35.2 & 30.2 \\
    Feat.~Rec.~(LI) &49.4 &  27.8& 79.8 & 28.4 & 35.3 & 31.4 \\
    \midrule
    \method~(RI) & 62.9 & 32.4 & 75.3 & 28.4 & 36.1 & 30.8 \\
    \rowcolor{TableColor} \method~(LI) & 69.3 &32.7 & 82.9 & 29.2 & 38.0 & 33.7 \\
    \bottomrule
\end{tabular}
}
\label{tab:init}
\end{table}

\begin{figure}[b]
    \centering
    \includegraphics[width=.9\columnwidth]{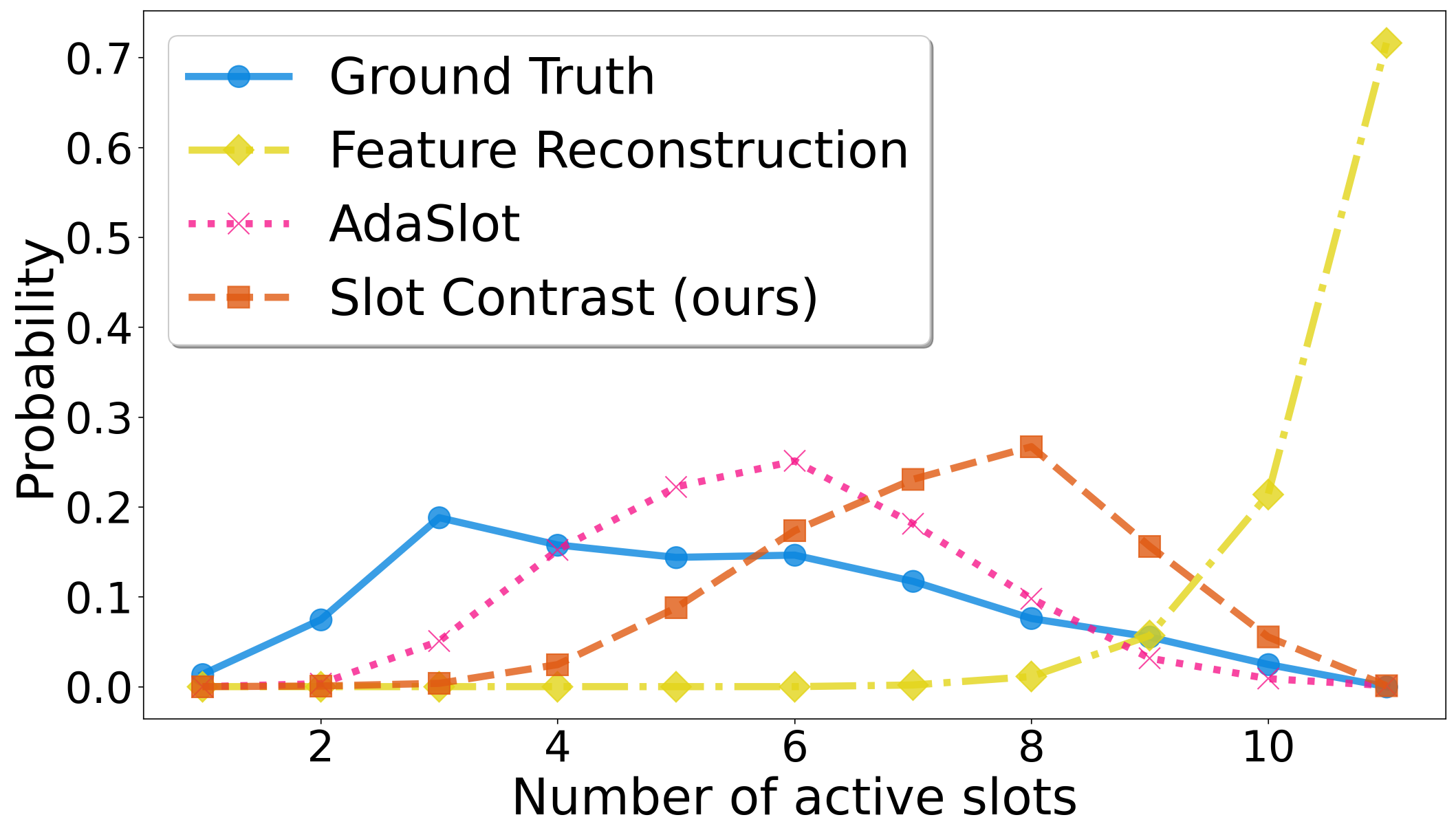}\vspace{-.8em}
    \caption{Distribution of ground truth and predicted object numbers (i.e., number of active slots) on the MOVi-C dataset. }
    \label{fig:number_slots}
\end{figure}
\paragraph{Choice of the First Frame Initialization (\Tab{tab:init})}
In video object-centric learning, slots are typically initialized based on those from the previous time step~\citep{kipf2022conditional}, while the first frame is initialized from learnable parameters.
% For the initial frame, slots can be initialized either randomly (e.g., using a Gaussian distribution with learned mean and variance) or via learned fixed initial states.
Previous real-world object-centric methods mostly used random initialization samples from Gaussian distribution~\citep{zadaianchuk2023objectcentric, aydemir2023self}, in this work, we study the impact of the type of initialization under our contrastive objective. The findings are outlined in \tab{tab:init}. Our experiments indicate that when combined with slot-slot contrastive loss, learned initialization significantly outperforms random initialization. We hypothesize that this improvement stems from the ability of learned initializations to shape the initial state in a way that enhances contrastiveness, a benefit not achievable with random initialization. %, where no such guidance exists.
For similarity visualizations and more details, see ~\app{app:learned_init}.

\paragraph{Number of Active Slots per Video~(\Fig{fig:number_slots})}

In models based on the Slot Attention mechanism, all available slots are typically utilized~\citep{seitzer2023bridging}, leading to a mismatch between the predicted number of components in scene decomposition and the ground-truth number of objects in the scene. This can cause the random splitting of the objects between slots and non-consistent scene representations when slots are reassigned from one object to a part of another object.
To address this challenge, it is important to study whether redundant slots can be effectively deactivated. Recently, AdaSlot~\citep{fan2024adaptive} introduced a discrete slot sampling module, coupled with a complexity-aware prior, to penalize redundant slots explicitly. Similarly, SOLV~\citep{aydemir2023self} used agglomerative clustering to merge redundant slots. In this work, we investigate whether \method is capable of accurately determining the number of objects in a scene without relying on explicit priors to minimize the number of active slots.
% In models based on the Slot Attention mechanism, all available slots are typically utilized~\citep{seitzer2023bridging}, leading to a mismatch between the predicted number of components in scene decomposition and the ground-truth number of objects in the scene. This can cause the random splitting of the objects between slots and non-consistent scene representations when slots are reassigned from one object to part of another object. Thus, for real-world video decomposition, it is essential to analyze whether redundant slots are effectively shut down, which becomes crucial. Recently, AdaSlot~\citep{fan2024adaptive} proposed to explicitly penalize redundant slots by a discrete slot sampling module with complexity-aware prior, while SOLV~\citep{aydemir2023self} proposed to merge redundant slots by postdoc agglomeration clustering procedure. Thus, we also study how good the \method is at automatically determining the number of objects while not having any explicit priors toward a minimal number of objects.

We compare ground truth and predicted object density on MOVI-C dataset, as shown in~\fig{fig:number_slots}. While the feature reconstruction model yields predictions within a narrow range---creating a sharp peak near a predefined number of slots---our model, similarly to AdaSlot~\citep{fan2024adaptive}, achieves a smoother prediction distribution that aligns more closely with the ground truth (note that the consistent shift is because 2--3 slots are used for the background, while the ground truth density is computed only for foreground objects). Interestingly, \method achieves this without requiring an explicit prior toward sparsity.\looseness=-1

\section{Conclusion}
\label{sec:conclusion}
\method advances unsupervised video object-centric learning by significantly improving the temporal consistency of object representations. 
Our method explicitly incentivizes temporal consistency by adding a self-supervised  contrastive loss. 
We showed that this loss is not only beneficial for consistency, but also enhances object discovery: \method achieves state-of-the-art results on challenging synthetic datasets with many objects and the unconstrained real-world YouTube-VIS dataset. 
Furthermore, consistent representations directly support temporal downstream tasks such as unsupervised object dynamics prediction and allow for tracking of objects through full occlusions. 
% Finally, \method effectively shuts down non-unique slots, making the representation more suitable for applications like autonomous control.
Finally, \method effectively shuts down non-unique slots, leading to a sparser representation that captures the true object distribution more faithfully.
Taken together, we expect these improvements to pave the way for broader adoption of video object-centric representations, for instance in applications like word modeling, autonomous control, or video question answering.

% Limitations should go here
%Our work has the following limitations: The number of slots is fixed as we learn initializations. A smaller number of slots can be used at test time, but we have not studied that. 
%Furthermore, there is no direct control over the granularity of the emerging segmentation of entities. 
Limitations of our work include the fixed number of slots during initialization. % and unexplored use of fewer slots at test time. 
Additionally, we cannot directly control the segmentation granularity of entities. Further limitations and failure cases are discussed in \app{app:limitations}.

Future work could explore several promising directions. First, one could use \method's robust and consistent representations for learning compositional world models from real-world robotics data to enable object-centric planning and control. Second, investigating the compatibility of our contrastive loss with other object-centric learning approaches with different inductive biases, such as  SlotDiffusion~\citep{wu2023slotdiffusion}. Finally, improving the compactness of object masks~\citep{kakogeorgiou2024spot} to achieve more precise object segmentation masks could also benefit unsupervised class-agnostic video object segmentation applications.
\section*{Acknowledgments}
We thank Christian Gumbsch for useful suggestions and Ke Fan for providing additional details and results for AdaSlot method. Andrii Zadaianchuk is funded by the European Union (ERC, EVA, 950086). The authors thank the International Max Planck Research School for Intelligent Systems (IMPRS-IS) for supporting Maximilian Seitzer. Georg Martius is a member of the Machine Learning Cluster of Excellence, EXC number 2064/1 – Project number 390727645.
This work was supported by the ERC - 101045454 REAL-RL. We acknowledge the support from the German Federal Ministry of Education and Research (BMBF) through the Tübingen AI Center (FKZ: 01IS18039B).
{
    \small
    \bibliographystyle{ieeenat_fullname}
    \bibliography{main}
}

% WARNING: do not forget to delete the supplementary pages from your submission, comment the next two lines and uncomment above the myexternal.... line
\clearpage
 % \setcounter{page}{1}
% Figures, Tables and Equations will have S in the name
\renewcommand{\thetable}{S\arabic{table}}
\renewcommand{\thefigure}{S\arabic{figure}}
\renewcommand{\theequation}{S\arabic{equation}}
\setcounter{table}{0}
\setcounter{figure}{0}
\setcounter{equation}{0}
% \resetlinenumber
\appendix

\maketitlesupplementary
\section{Training Details}  
 The general hyperparameters utilized during training \method are outlined in \tab{tab:implementation_details}, ensuring clarity and reproducibility. Furthermore, the task-specific hyperparameters used for object dynamics prediction are detailed separately in \tab{tab:implementation_details_sf}.
\label{app:training_details}
\begin{table*}[h]
    \centering
    \caption{Hyperparameters of Slot-Slot Contrast Model for Main Results on MOVi-C, MOVi-E, and YouTube-VIS 2021 Datasets}
    \begin{tabular}{@{}lccc@{}}
        \toprule
        \textbf{Hyperparameter} & \textbf{MOVi-C} & \textbf{MOVi-E} & \textbf{YouTube-VIS} \\
        \midrule
        Training Steps & 100k & 300k & 100k \\
        Batch Size & 64 & 64 & 64 \\
        Training Segment Length & 4 & 4 & 4 \\
        Learning Rate Warmup Steps & 2500 & 2500 & 2500 \\
        Optimizer & Adam & Adam & Adam \\
        Peak Learning Rate & 0.0004 & 0.0008 & 0.0008 \\
        Exponential Decay & 100k & 300k & 100k \\
        ViT Architecture & DINOv2 Small & DINOv2 Base & DINOv2 Base \\
        Initialization & FixedLearnedInit & FixedLearnedInit & FixedLearnedInit \\
        Patch Size & 14 & 14 & 14 \\
        Feature Dimension ($D_{\text{feat}}$) & 384 & 768 & 768 \\
        Gradient Norm Clipping & 0.05 & 0.05 & 0.05 \\
        \midrule
        \textbf{Image Specifications} & & & \\
        Image / Crop Size & 336 & 336 & 518 \\
        Cropping Strategy & Full & Full & Rand. Center Crop \\
        Augmentations & -- & -- & Rand. Horizontal Flip \\
        Image Tokens & 576 & 576 & 1369 \\
        \midrule
        \textbf{Slot Attention} & & & \\
        Slots & 11 & 15 & 7 \\
        Iterations (first / other frames) & 3 / 2 & 3 / 2 & 3 / 2 \\
        Slot Dimension ($D_{\text{slots}}$) & 64 & 128 & 64 \\
        \midrule
        \textbf{Predictor}  & & & \\
        Type & Transformer& Transformer& Transformer \\
        Layers & 1 & 1 & 1 \\
        Heads & 4 & 4 & 4 \\ 
        \midrule
        \textbf{Decoder} & & & \\
        Type & MLP & MLP & MLP \\
        \midrule
        \textbf{Loss Parameters} & & & \\
        Softmax Temperature ($\tau$) & 0.1 & 0.1 & 0.1 \\
        Slot-Slot Contrast Weight ($\alpha$) & 0.5 & 1 & 0.5 \\
        \bottomrule
    \label{tab:implementation_details}
    \end{tabular}
\end{table*}

\section{Effect of Learned Initialization}
\begin{figure*}[ht]
    \centering
    \begin{subfigure}{0.49\textwidth}
        \centering
        \includegraphics[width=\textwidth]{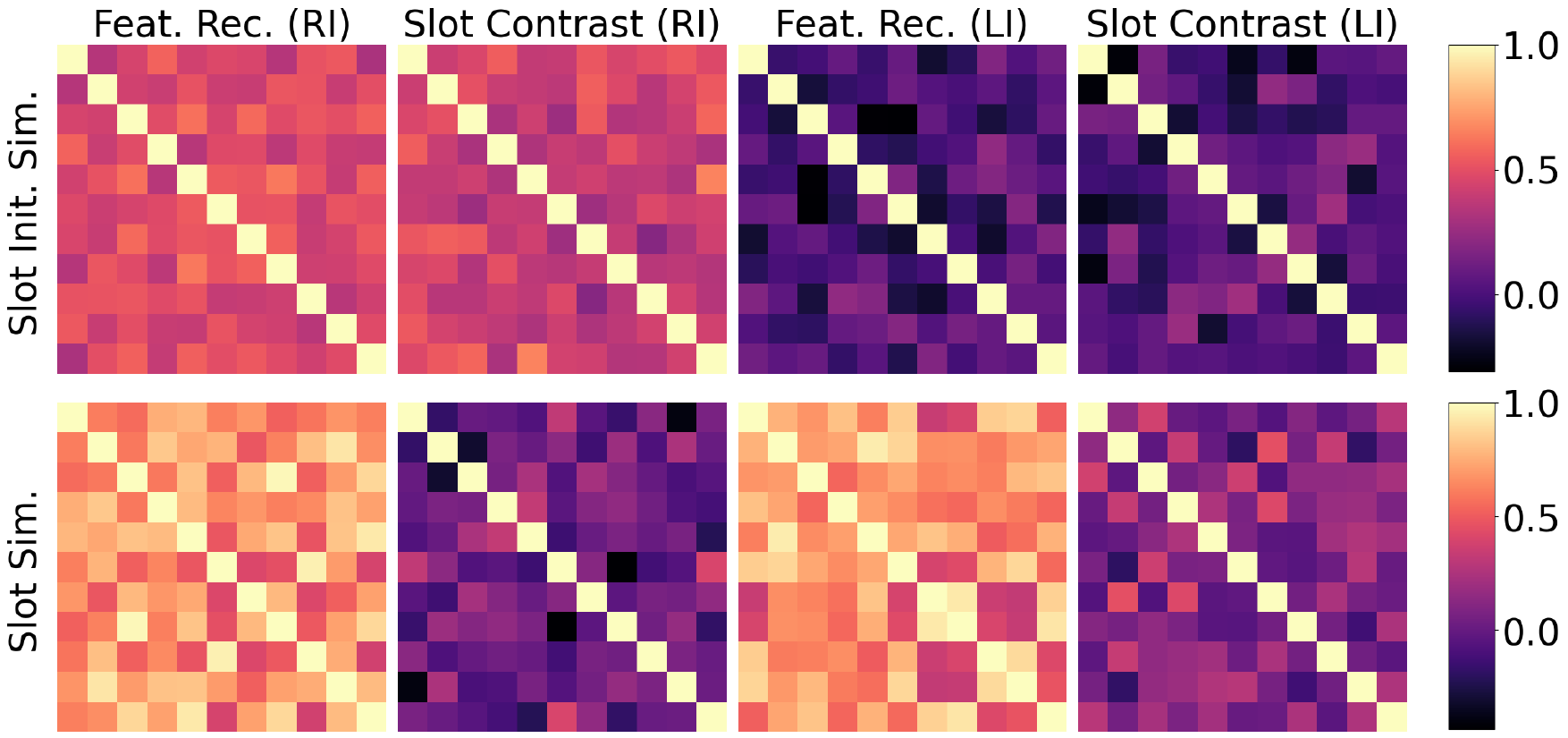}
        \caption{MOVi-C dataset}
        \label{fig:init_sim_movi_c}
    \end{subfigure}
    \hfill
    \begin{subfigure}{0.49\textwidth}
        \centering
        \includegraphics[width=\textwidth]{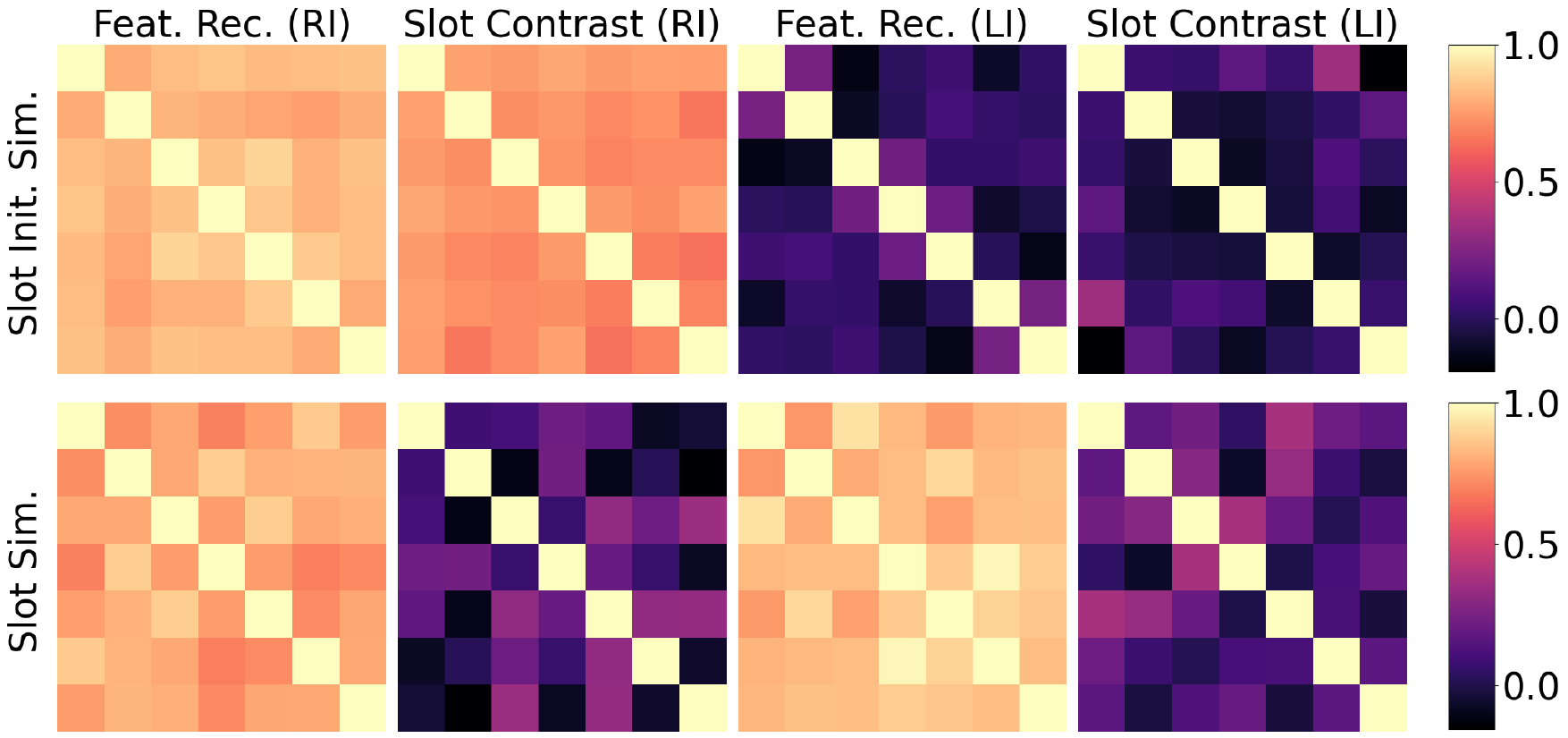}
        \caption{YT-VIS dataset}
        \label{fig:init_sim_ytvis}
    \end{subfigure}
    \caption{Similarity matrix between the set of slot initializations, $S_0$ (first row) and first frame slots, $S_1$ (second row) for different loss functions (feature reconstruction and slot-slot contrast loss) and different initialization strategies (RI = random initialization; LI = learned initialization).}
    \label{fig:learned_init_combined}
\end{figure*}

To determine the optimal approach for first-frame slot initialization, we compared two techniques: sampling from a random distribution and learning fixed query vectors. Our experimental results show that learned initialization consistently yields superior performance. We hypothesize that this improvement arises from the emergence of contrastive slots during learning, a desirable property that promotes slot specialization.
To illustrate this point, we visualized slot similarities for models initialized using both random and learned methods on the MOVi-C and YTVIS datasets (see the first row of \fig{fig:learned_init_combined}). The plots demonstrate a clear pattern: learned slot initializations produce more contrastive representations, highlighting their advantage over random initialization. In addition, using slot-slot contrastive loss, we maintain the constructiveness of the slots (see the second row of \fig{fig:learned_init_combined}), thus allowing for similar initialization for successive frame processing. 

Next, we further analyze possible slot initializations that are more flexible than fixed initialization but are still contrastive. In particular, we propose an additional adaptive initialization method using $k$-means clustering. In particular, we use $k$-means clustering on dense object-centric features $h_{0}$ obtained by adapting original patch DINO features with a simple MLP module $g_{\psi}$. The cluster centroids (that are naturally not similar to each other) serve as slot initialization for the initial frame in the video. \method trained with such adaptive initialization achieves an FG-ARI score of $73.1$ on the MOVi-C dataset ($+2.8$ FG-ARI improvement from fixed initialization). This result highlights the importance of flexible and contrastive first-frame slot initialization on model performance. However, the adaptive initialization is not scalable due to the significant computational overhead of running $k$-means for each initialization. Despite this limitation, the proof of concept demonstrates the promise of advanced initialization strategies, inviting further research in this direction.

\label{app:learned_init}

\section{Implementation of Slot-Slot Contrastive Loss}
\label{app:loss_details}
In this section, we provide details on the practical implementation of the slot-slot contrastive loss.
Given the slot representations $s_t$ and $ s_{t+1} $ at time steps $ t $ and $ t+1 $, we compute the similarity matrix $\mathbf{A}$:
\begin{equation}
A^{ij}_{t,t+1} = \frac{s^{i}_{t} \cdot s^{j}_{t+1}}{\|s^{i}_{t}\| \|s^{j}_{t+1}\|}    
\end{equation}
where each element $ A^{ij}_{t,t+1} $ represents cosine similarity between the $ i $-th slot at time $ t $ and the $ j $-th slot at time $ t+1 $.

Next, we apply the cross-entropy loss $\mathcal{L}_{\text{CE}}(\mathbf{P}, \mathbf{I})$ between the computed softmax normalized slot similarities $\mathbf{P} = \mathrm{softmax}{(\mathbf{A})} $ and the identity matrix $\mathbf{I}$.
\paragraph{Batch Contrastive Loss} We modify the similarity matrix $\mathbf{A}$ to include not only the slots for the current frame at time step $t$ and the subsequent frame at time step $t+1$, but also the slots from all frames within the batch of videos that are processed together. 
Let $B$, $T$, $K$, and $D$ denote the batch size, sequence length, number of slots, and the dimension of the slots, respectively. 
Initially, the similarity matrix $\mathbf{A}$ has shape $\mathbf{A} \in \mathbb{R}^{B \times (T-1) \times K \times K}$. 
After modifying it for batch comparison, its shape becomes $\mathbf{A}' \in \mathbb{R}^{(T-1) \times (K B) \times (K B)}$. 

\section{Feature Reconstruction Loss as Regularizer}
\label{app:feat_rec}
To promote better object discovery we also use feature reconstruction loss. Feature reconstruction loss, $\mathcal{L}_{\text{rec}}$, measures the discrepancy between the predicted features $\hat{h}_t$ and the true features $h_t$ at each time step $t$. In our case the features correspond to self-supervised DINOv2 features. 
The loss could be computed using a common distance metric such as Mean Squared Error (MSE):

\begin{equation}
\mathcal{L}_{\text{rec}} = \sum_{t=1}^{T-1} ||h_t - \hat{h}_t||^2    
\end{equation}

The loss also serves as an effective regularizer, mitigating undesired behaviors that can arise from the contrastive nature of slot-slot contrastive loss.
For example,  slot-slot contrastiveloss can't pull slots representing different objects together because it is minimized alongside the feature reconstruction loss $\mathcal{L}_{\text{rec}}$. This way, we maximize slot-slot similarity while still requiring each slot to be informative about original inputs. So \emph{region-wise reconstruction} with an MLP decoder decoding slots individually is an \emph{effective regularizer}, preventing ``wrong slots pulling'' behavior as otherwise pulled slots will not contain the information about the object they are responsible to reconstruct. 

Another key scenario is when an object disappears. In this case, it is important to understand what happens to the corresponding slot and how its behavior is governed by the objectives. In that case, we want the corresponding slot to maintain object information. 
 Given the additional reconstruction loss, it is possible by ignoring the disappeared object’s slot (thus serving as latent memory until object reappearance).
This behavior is evident in the \fig{fig:number_slots} showing \emph{fewer active slots} compared to baseline that uses all the available slots.\looseness=-1 

\section{Dataset Details}
\label{app:dataset_details}
In this section, we provide details about the datasets used in our work. Overall, we use several synthetic datasets (MOVi-C and MOVi-E) and one challenging real-world dataset, YouTube-VIS. For all datasets, annotations are used only during the evaluation of the object discovery, while during training, we use only videos from the datasets.
\paragraph{MOVi Datasets} 
 For both MOVi-C and MOVi-E, we utilized the standard train/validation splits. Each dataset contains $9750$ training sequences and $250$ validation sequences. While the original datasets are provided at a resolution of $256\times256$, we resized them to $336\times336$ for our experiments. It is important to note that we did not generate new datasets, but rather modified the resolution of the original data. This way, we make sure that all the methods are comparable in terms of both original input resolution while using a similar or less token during ViT processing ($576$ for \method and VideoSAURv2, and $784$ tokens for original VideoSAUR~\citep{zadaianchuk2023objectcentric}).

\paragraph{Youtube-VIS 2021} 
The YouTube-VIS dataset is an unconstrained, real-world dataset designed for video instance segmentation. It has two versions: YouTube-VIS 2019 and YouTube-VIS 2021. In our work, we used YouTube-VIS 2021, as it is more complex and challenging compared to the 2019 version.
We split the original training set into a new training set and a validation set, comprising 2,775 and 210 videos, respectively. This split was necessary because the original validation set for YouTube-VIS 2021 is not publicly available.

\section{Metrics Details}
\label{app:metrics_details}
To evaluate our method, we use two metrics: foreground Adjusted Rand Index (FG-ARI) and mean Best Overlap (mBO) to assess the quality of the masks produced by our models. FG-ARI is a variant of the standard ARI metric, computed by excluding the background mask, and is commonly used in the object-centric literature to measure the similarity between predicted object masks and ground truth masks. It primarily evaluates how well objects are segmented.

Mean Best Overlap (mBO), on the other hand, measures the similarity between predicted and ground truth masks using the intersection-over-union (IoU). For each ground truth mask, the predicted mask with the highest IoU is selected, and the average IoU is computed across all matched pairs. mBO also considers background pixels, offering a better measure of how well the masks align with the objects.

To differentiate between per-frame (image-based) and video-wide evaluations, we use "Image" as a prefix for the metrics (e.g., Image FG-ARI and Image mBO) when computed on individual frames.  When we do not use an additional prefix, we refer to the "Video" version of the same metric when computed across entire videos. We are particularly interested in video-based metrics, as they additionally consider the consistency of object masks.

\section{Baseline Details}
\label{app:baseline_details}
\paragraph{VideoSAUR}
To compare our method with the state-of-the-art VideoSAUR method~\citep{zadaianchuk2023objectcentric}, we considered two configurations: VideoSAUR trained with DINO features~\citep{Caron2021DINO} and VideoSAUR trained with DINOv2~\citep{oquab2023dinov2} features, which we refer to as VideoSAURv2. 

For the YouTube-VIS 2021 dataset, the authors of VideoSAUR provided results for both configurations, so we directly used the available checkpoints. However, for the MOVi datasets, results and model for VideoSAUR trained with DINOv2 features were not available. Therefore, we trained VideoSAUR with the default configuration( matching the resolution with \method) using DINOv2 features. 

While for MOVI-E the default configuration with DINOv2 lead to improved results, MOVi-C results were significantly worse.  Thus,  we perform an extensive hyperparameter tuning, experimenting with the weight of the temporal similarity loss, temperature parameters, with and without feature reconstruction loss added. We also tested various configurations of keys, values, and output features from the Vision Transformer. Despite these efforts, we could not achieve performance comparable or better to VideoSAUR trained with DINOv1 features. Our best performing VideoSAURv2 configuration ($62.1$ FG-ARI and $25.5$ mBO) on MOVi-C is obtained using temperature $\tau=0.075$ temporal similarity loss weight $\alpha=0.1$ combined with feature reconstruction loss. We also used DINOv2 ViT \textit{values} features in contrast to \textit{keys} features used in the original VideoSAUR paper~\citep{zadaianchuk2023objectcentric} with DINOv1.

This discrepancy raises the question: why does VideoSAURv2 work well on MOVi-E and YouTube-VIS but not on simpler MOVi-C? We hypothesize that the presence of camera motion in MOVi-E might contribute to the success of DINOv2 features in this context. To test this hypothesis, one can evaluate VideoSAUR on the MOVi-D dataset, which is similar in complexity to MOVi-E, but lacks camera motion.\looseness=-1

\paragraph{SAM2}
To compare how close current object-centric methods are to supervised methods  we compared \method with SAM2 as a supervised zero-shot baseline for temporal consistency. 
As SAM2 is trained on a large dataset with dense video annotations~($190.9K$ masklets), using its tracking can improve segmentation consistency~(limited to objects discovered in the first frame). However, while SAM2 can be used only for object tracking, \emph{our method is not limited to tracking}; it jointly does both object discovery in videos and learns consistent object representations with their masks. We evaluate SAM2's tracking capabilities by combining SAM2 with initial frame object discovery using video-based DINOSAUR~(i.e, feature reconstruction objective on videos) and \method object discovery~(see~\tab{tab:sam}). We show that \method halves the gap between unsupervised object-centric learning and zero-shot SAM2~($5.5$ vs $12.3$ FG-ARI), while using \method object discovery is helpful for overall tracking with SAM2 ($+2.8$ FG-ARI).\looseness=-1
\begin{table}
\centering
    \caption{Temporal consistency on YouTube-VIS 2021.}
    \label{tab:movi_youtube_metrics_supp}
    \normalsize
    \setlength{\tabcolsep}{3pt} 
    \adjustbox{width=\linewidth}{
   \begin{tabular}{ccccc} 
    \toprule
       & \textbf{Feat. Rec. + SAM2} & \textbf{\method + SAM2} & \textbf{VideoSAURv2} & \textbf{\method} \\
    \midrule
    \textbf{FG-ARI}  & 43.5 & 46.3  & 31.2 & 38.0 \\
    \textbf{mBO}     & 40.9 & 43.7  & 29.7 & 33.7 \\
    \bottomrule
\end{tabular}}
\label{tab:sam}
\end{table}

\begin{figure}[ht]
    \centering
    \hspace{-1.4em}
    \includegraphics[width=0.4\textwidth]{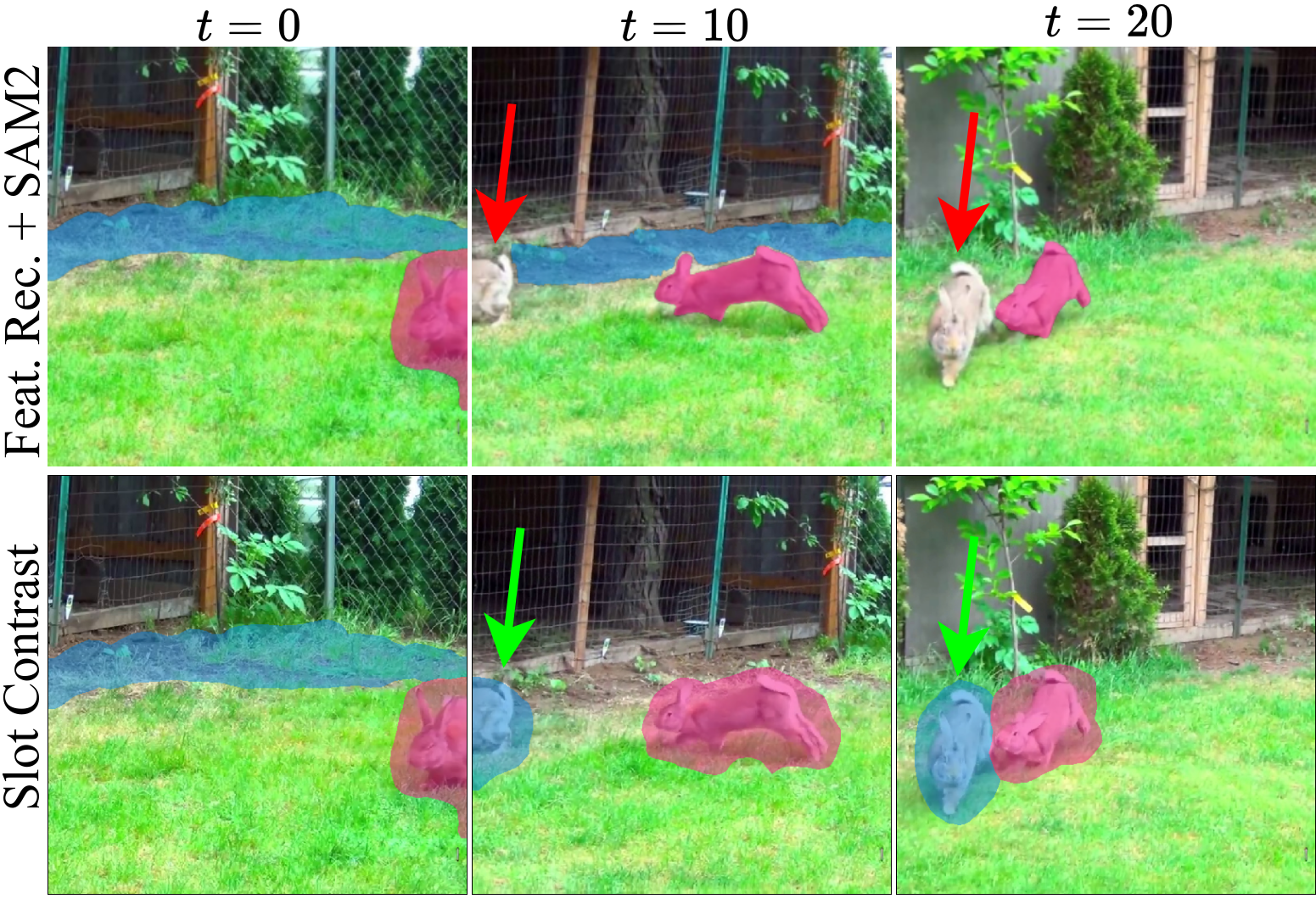}
    \caption{SlotContrast vs SAM2 tracking. SAM2 is limited to track only objects that appeared and discovered in the first frame.\looseness=-1}
    \label{fig:sam_vis}
\end{figure}
In addition, in \fig{fig:sam_vis}, we show limitation of such baseline: detecting and tracking later appearing objects due to missing initial masks. Evaluating SAM2 on YTVIS's first-frame objects gives $46.3$ mBO ($+6\%$), while for the later-appearing objects, mBO drops to $7.82$ ($-34.48\%$).\looseness=-1 
This highlights SAM2's strength in tracking first-frame objects and its limitation in detecting and tracking later objects due to missing initial masks.
\paragraph{SAVI++}
\label{savi++}
We compared \method with weakly supervised method SAVi++. We used improved SAVi similar to VideoSAUR (see App. C.5 VideoSAUR), \emph{reaching $42.8$ FG-ARI on MOVi-E}. In contrast,  unconditioned optical-flow SAVi and depth SAVi++ are only $28.1$ and $31.7$ as reported by \citet{bao2023object}. While adding depth signal in \textbf{SAVi++} could be treated as weak supervision, it indeed improves SAVi $16.0$ mBO, reaching $22.1$ mBO, but \emph{still lagging behind both VideoSAUR and SlotContrast}.

\section{Per-frame Scene Decomposition}
\label{app:image}
In this section, we extend our comparison for the scene decomposition task to the MOVi-C dataset. The results are presented in \tab{tab:image_extended}. Our method outperforms all state-of-the-art approaches by a significant margin, with the sole exception of VideoSAUR, where we observe a minor performance gap of just 0.4 points, indicating comparable results. 
\begin{table}[H]
\centering
\begin{adjustbox}{max width=\columnwidth}
\begin{tabular}{l l l c} % Adjusted column layout
    \toprule
    \multirow{2}{*}{} & \textbf{Model} & \textbf{Objective} & \multicolumn{1}{c}{\textbf{Image}} \\
    % \cmidrule(lr){4-4}
    &&& \textbf{FG-ARI} \\
    \midrule
    \multirow{3}{*}{$\mathcal{I}$}
    & LSD~\citep{jiang2023object} & Image Rec. &  50.5 \\
    & DINOSAUR~\citep{seitzer2023bridging} & Image Rec. &  68.6 \\
    \midrule
    $\mathcal{V}+\mathcal{M}$ & Safadoust et al.~\citep{safadoust2023multi} & +GT Flow & 73.8 \\
    \midrule
    \multirow{4}{*}{$\mathcal{V}$} 
    & STEVE~\citep{Singh2022STEVE} & Video Rec. &  51.9 \\
    & VideoSAUR~\citep{zadaianchuk2023objectcentric} & Temp. Sim. & \textbf{75.5} \\
    % & VideoSAURv2~\citep{zadaianchuk2023objectcentric} & Temp. Sim. & - \\
    & Feat. Rec. & Video Rec. & 64.0 \\
    & \cellcolor{TableColor}\method & \cellcolor{TableColor}Slot Contrast & \cellcolor{TableColor}75.1 \\
    \bottomrule
\end{tabular}
\end{adjustbox}
\caption{Quantitative Results on MOVi-C dataset in terms of per-frame Image FG-ARI. The methods are grouped by the target data they train on: only images ($\mathcal{I}$), videos with motion segmentation annotations ($\mathcal{V} + \mathcal{M}$), and only videos ($\mathcal{V}$).}
\label{tab:image_extended}
\end{table}
Finally, on the YTVIS dataset for the image decomposition task, our method achieves a FG-ARI of $45.1$ outperforming both VideoSAUR~($40.1$ FG-ARI) and VideoSAURv2~($40.5$ FG-ARI). 
% Additionally, Feature Reconstruction achieves a score of $44.1$, demonstrating that our method performs slightly better.

\section{Instance-Awareness of Dense Features}
\label{app:learned_features}
In this section, we emphasize the need to adapt self-supervised DINOv2 ViT features for consistent object discovery. While DINOv2 features are primarily semantic, they need refinement to identify specific instances effectively. To facilitate this, we project the frozen features through a multi-layer perceptron (MLP). This transformation maps the features into a new latent space, enhancing their instance-awareness and simplifying the Slot Attention task. 

\begin{figure}[ht]
    \centering
    \includegraphics[width=0.4\textwidth]{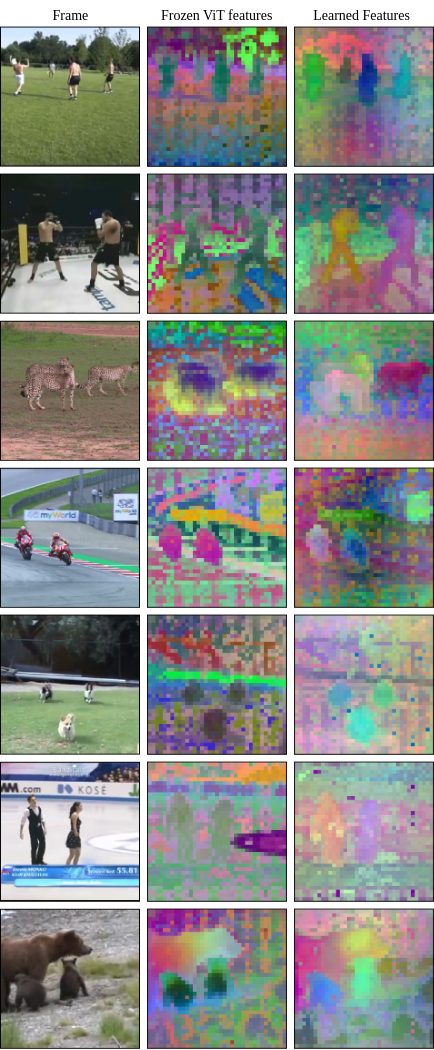}
    \caption{First three Principal Components (combined as RGB channels into one image for convenience) of frozen DINOv2 features and the newly learned dense features. DINOv2 features PCA components are semantic grouping instances of the same category (e.g., people or dogs) and body parts of the different instances~(e.g., heads or legs). In contrast, learned dense features have instance-aware components, separating different instances of the same category, thus making object discovery easier.\looseness=-1}
    \label{fig:pca}
\end{figure}

To show the effect of this adaptation on dense features, we visualize the first Principal Component Analysis (PCA) of both the frozen DINOv2 features and the newly learned adapted dense features~(see the results in \fig{fig:pca}). The PCA plots clearly show that while DINO features cluster similarly across different instances, the learned features are more distinct, effectively capturing instance-specific details.

Further, we evaluate the effectiveness of these instance-aware features by conducting experiments with both frozen and learned features. The results, summarized in \tab{tab:dense_feat}. While MOVi-C, where most of the time different objects have different semantic categories, adapting shows minor improvement, the improvements are substantial for MOVi-E and the real-world YouTube-VIS dataset. This demonstrates the clear advantage of learning to adapt DINOv2 features to be instance-aware in challenging real-world scenarios.

\begin{table}[tb]
\centering
    \caption{Comparison of consistent object discovery evaluated by Video FG-ARI. We compare \method with frozen DINOv2 features and \method based on additionally adapted with MLP dense features.}
    % \vspace{-0.5em}
    \setlength{\tabcolsep}{2.2pt} % Reduce the space between columns
    \adjustbox{max width=\linewidth}{
   \begin{tabular}{l@{}ccc} % Adjust column layout
    \toprule
       & \multicolumn{1}{c}{\textbf{MOVi-C}} & \multicolumn{1}{c}{\textbf{MOVi-E}} & \multicolumn{1}{c@{}}{\textbf{YouTube-VIS}} \\
    % \cmidrule(lr){2-2} \cmidrule(lr){3-3} \cmidrule(lr){4-4}
    % & \textbf{FG-ARI} $\uparrow$ & \textbf{FG-ARI} $\uparrow$ & \textbf{FG-ARI} $\uparrow$ \\
    \midrule
    Frozen DINOv2 Features & 68.4 & 75.3 & 33.7 \\
    \rowcolor{TableColor}MLP Adapted Features & 69.3 & 82.9 & 38.0 \\
    \bottomrule
\end{tabular}
}
    \label{tab:dense_feat}
%\vspace{-2em}
\end{table}

\section{SlotFormer}
\label{app:slotformer}

To evaluate our model's performance on the object dynamics prediction task, we trained a SlotFormer~\citep{wu2023slotformer} module on top of our object-centric model. The code for SlotFormer was taken from its official codebase\footnote{\url{https://github.com/pairlab/SlotFormer}}. SlotFormer consists of a transformer encoder with input and output projection, and it adds positional embeddings to the input along the temporal dimension. It takes the slots from $T$ burn-in frames and then predicts the slots for the next $K$ rollout frames in an autoregressive manner. The model is trained by minimizing the mean squared error between the predicted slots and the ground-truth slots provided by the grouper. During training, the entire architecture of the object-centric model is frozen, and only the dynamics predictor module is optimized.  

The hyperparameters used for training the models are listed in \tab{tab:implementation_details_sf}. For MOVi-C, we used entire videos for both training and validation, with the first fourteen frames serving as burn-in frames, while the model predicted the slots for the remaining frames. MOVi-E videos are also 24 frames long, but we chose to evaluate performance on the middle segment of the video because most objects remain static in the final frames. To create a more challenging evaluation, we selected the first 5 frames as burn-in and predicted the slots for the next 10 frames. Finally, for YTVIS, we used the first 10 frames as burn-in and had the model predict only the following 5 frames due to the dataset's complexity.

\begin{table}[ht!]
    \centering
    \caption{Hyperparameters of SlotFormer for Main Results on MOVi-C, MOVi-E, and YouTube-VIS 2021 Datasets}
    \begin{adjustbox}{max width=\columnwidth}
    \begin{tabular}{@{}lccc@{}}
        \toprule
        \textbf{Hyperparameter} & \textbf{MOVi-C} & \textbf{MOVi-E} & \textbf{YouTube-VIS} \\
        \midrule
        Training Steps & 100k & 100k & 100k \\
        Batch Size & 128 & 128 & 128 \\
        Burn-in Steps $T$ & 14 & 5 & 10 \\
        Rollout Steps $K$ & 10 & 10 & 5 \\
        Latent Size $D_e$ & 128 & 256 & 128 \\
        Hidden Size of FFN & 512 & 1024 & 512 \\
        Number of Layers $N_{ \tau }$ & 1 & 1 & 4 \\
        Dropout Rate & 0.2 & 0.1 & 0.1 \\
        Peak Learning Rate & $2 \times 10^{-4}$ & $2 \times 10^{-5}$ & $10^{-5}$ \\
        \bottomrule
    \label{tab:implementation_details_sf}
    \end{tabular}
    \end{adjustbox}
\end{table}

\section{Details and Visual Examples on MOVI-C Occluded}
\label{app:movi-c-occluded visualizations}
We created a targeted subset of the MOVi-C dataset that focuses exclusively on fully occluded object sequences.
The MOVi-C dataset provides visibility scores for each object in each frame, indicating the number of pixels the object occupies. Using these scores, we refine the validation set to include only sequences meeting the following conditions: an object initially appears with a visibility score of at least $\mathrm{n}$ pixels, then becomes fully occluded (visibility score drops to 0 pixels), and subsequently reappears with a visibility score of at least $\mathrm{n}$ pixels.
To avoid including very small objects or visual artifacts, we set $\mathrm{n}$ to a minimum of $400$ pixels (less than 1\% of the image pixels).
After applying this filtering criterion, we obtain a dataset of 60 sequences where objects undergo complete occlusion and reappearance. Visualizations are presented in \fig{fig:occluded_subset}.

\section{Limitations and Failure Cases}
\label{app:limitations}
While \method demonstrates significant improvements over previous approaches, several limitations remain. One key area for improvement is the sharpness of predicted object masks, which could be tighter and sometimes occupy some background parts (referred to as ``bleeding'' artifacts). Another major challenge lies in ensuring consistency during long-term full occlusions. Although \method often reidentifies objects after such occlusions successfully, some failure cases persist. 

Additionally, \method lacks control over slot behavior when objects disappear. Ideally, slots corresponding to disappeared objects should remain inactive and not be decoded, but the current implementation leaves this decision to the decoder. Future work could address this by making the behavior more explicit. Lastly, \method relies on a predefined, fixed number of slots, which may limit its flexibility. We visualize some of the failure cases in \fig{fig:fail_examples}.

\section{Additional Examples}
\label{app:additional_examples}

In this section we present the following additional visualizations.

\begin{figure*}[ht]
    \centering
    \includegraphics[width=1\textwidth]{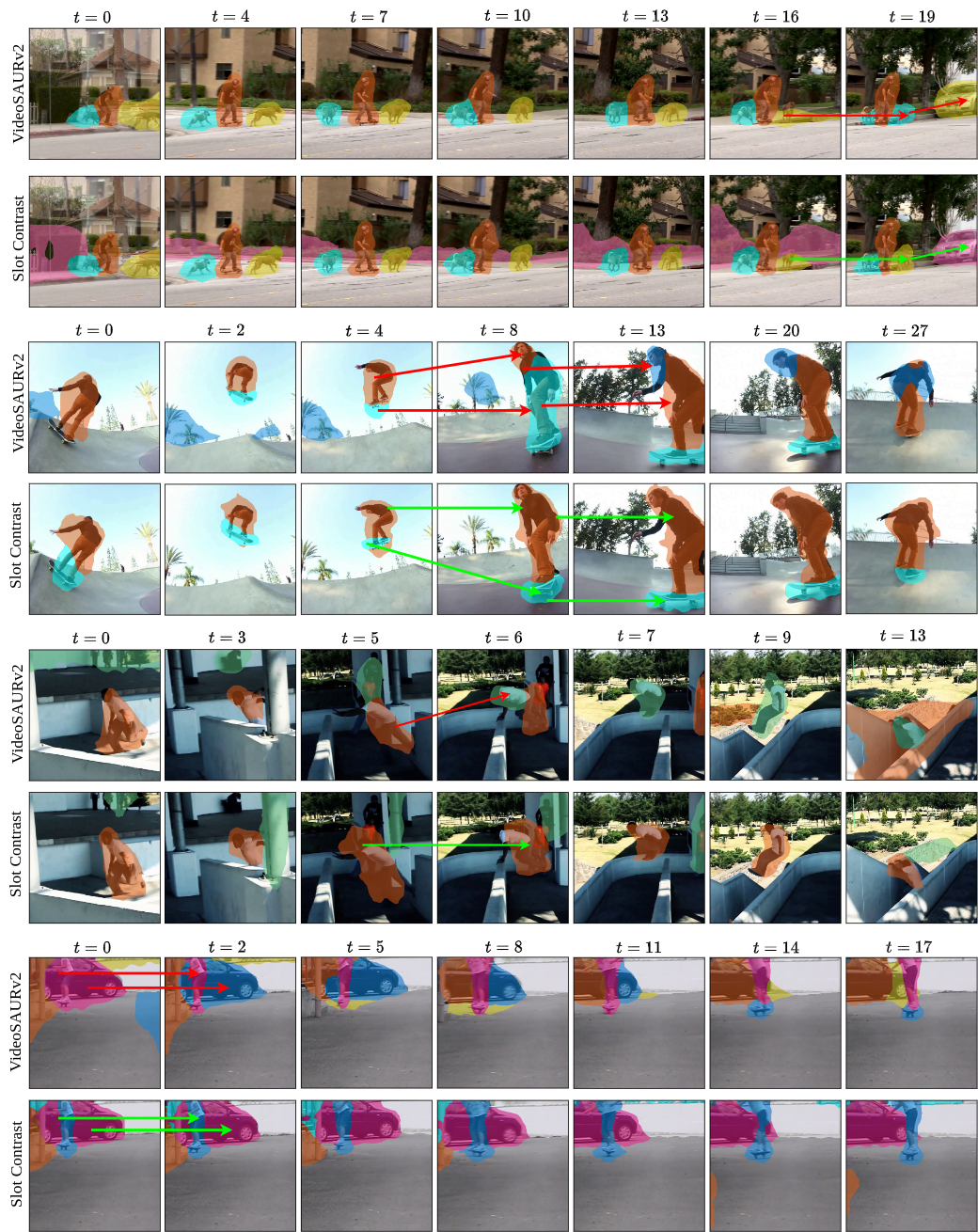}
    \caption{Qualitative comparison of \method with VideoSAURv2 on YouTube-VIS 2021 dataset.}
    \label{fig:videosaur_comparison}
\end{figure*}

\begin{figure*}[ht]
    \centering
    \includegraphics[width=1\textwidth]{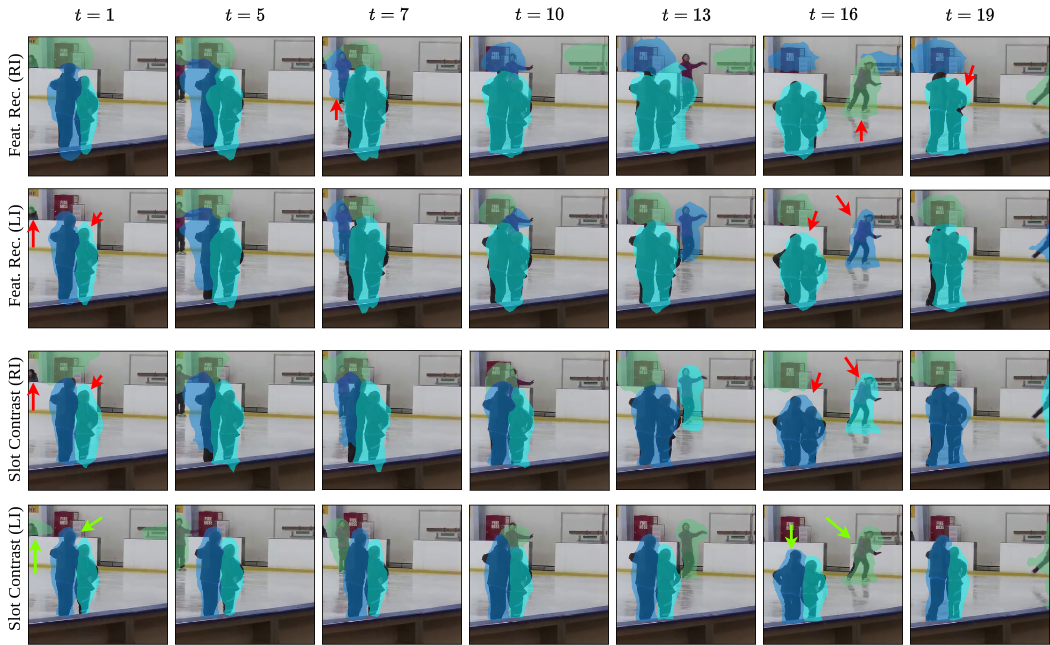}
    \caption{Qualitative results of first frame slot initialization ablations on YouTube-VIS 2021 dataset.}
    \label{fig:ablations_183}
\end{figure*}

\begin{figure*}[ht]
    \centering
    \includegraphics[width=1\textwidth]{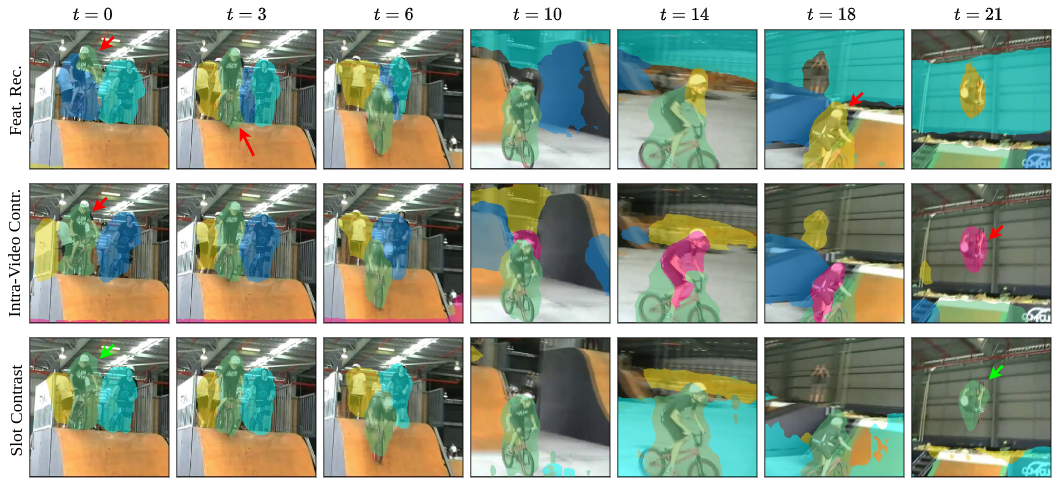}
    \caption{Qualitative results of loss function ablations on YouTube-VIS 2021 dataset. }
    \label{fig:ablations_148}
\end{figure*}

\begin{figure*}[ht]
    \centering
    \includegraphics[width=1\textwidth]{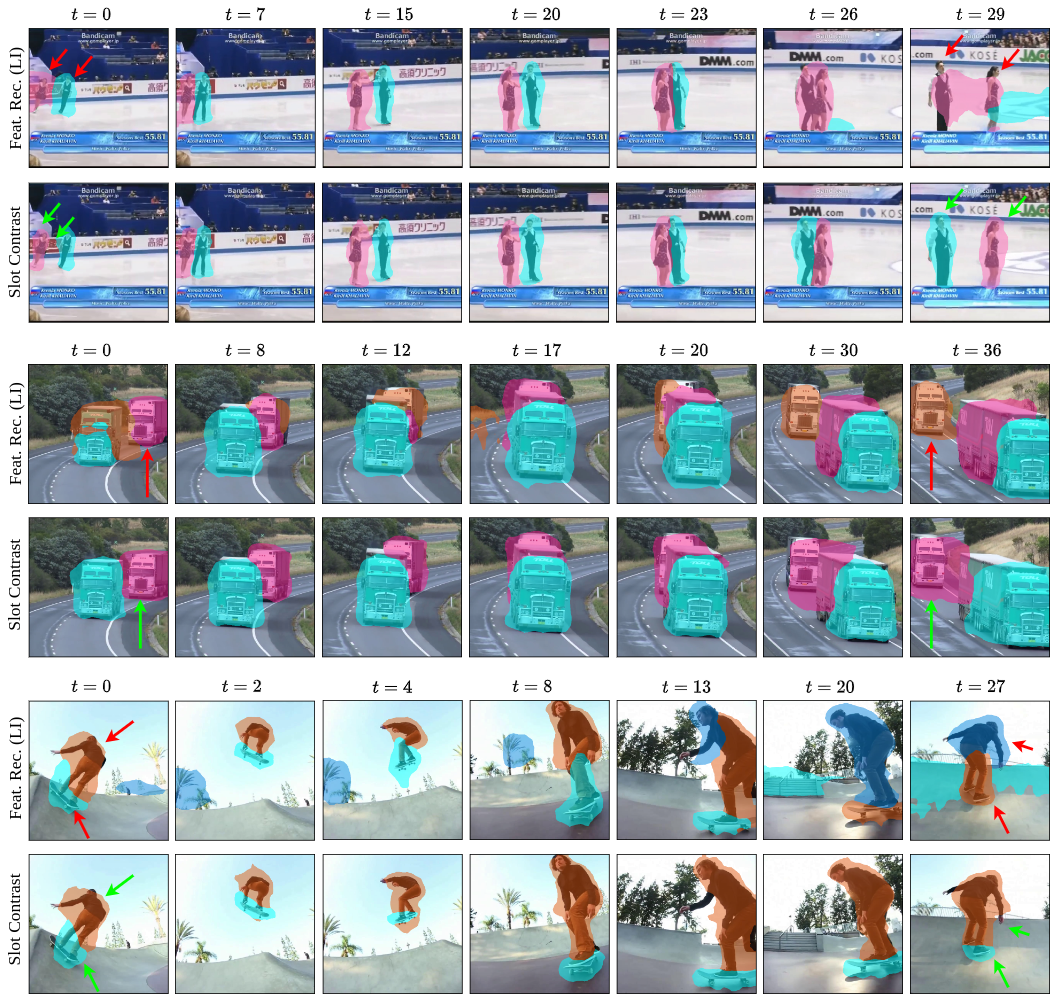}
    \caption{Qualitative comparison of \method with Features Reconstruction baseline with learned initialization on YouTube-VIS 2021 dataset.}
    \label{fig:ablations_feat_rec}
\end{figure*}

\begin{figure*}[ht]
    \centering
    \includegraphics[width=0.9\textwidth]{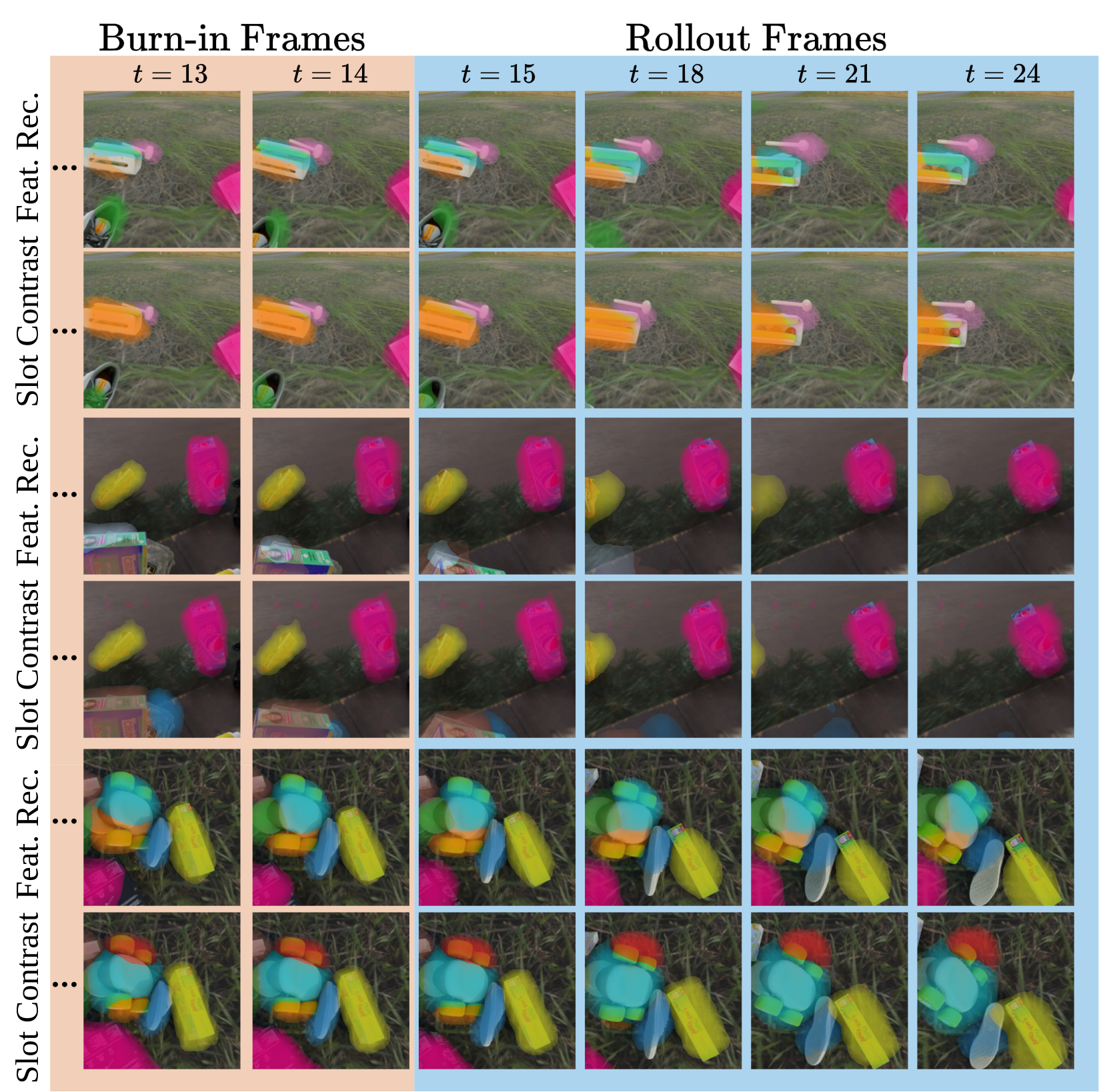}
    \caption{Comparison of masks obtained by decoding the predicted slots from SlotFormer, trained on top of the feature reconstruction baseline, versus \method, tested on the MOVi-C dataset.}
    \label{fig:slotformer_movic}
\end{figure*}

\begin{figure*}[ht]
    \centering
    \includegraphics[width=0.9\textwidth]{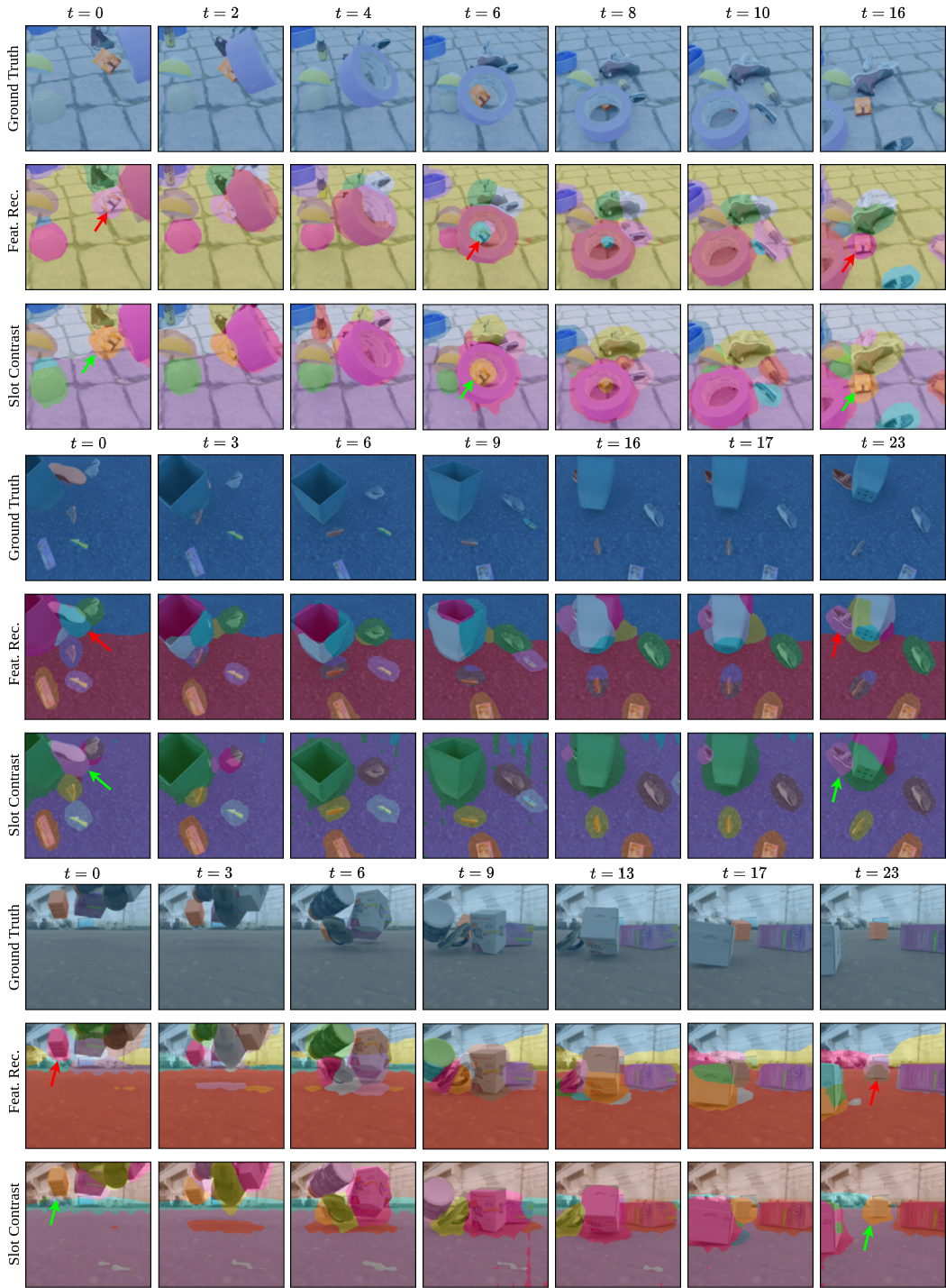}
    \caption{Qualitative comparison of \method with Features Reconstruction on MOVi-C occluded subset.}
    \label{fig:occluded_subset}
\end{figure*}

\begin{figure*}[ht]
    \centering
    \includegraphics[width=1\textwidth]{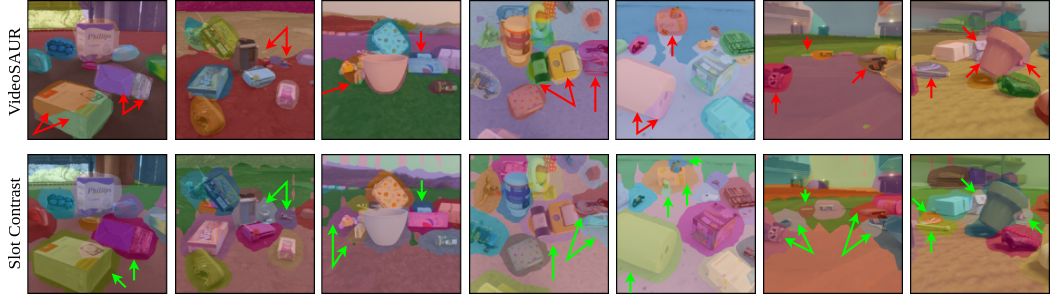}
    \caption{Example frames comparing \method and VideoSAUR on the MOVi-E scene decomposition task. VideoSAUR occasionally misses objects or splits one object into multiple slots, while these errors are avoided by \method.}
    \label{fig:movi_e}
\end{figure*}

\begin{figure*}[ht]
    \centering
    \includegraphics[width=1\textwidth]{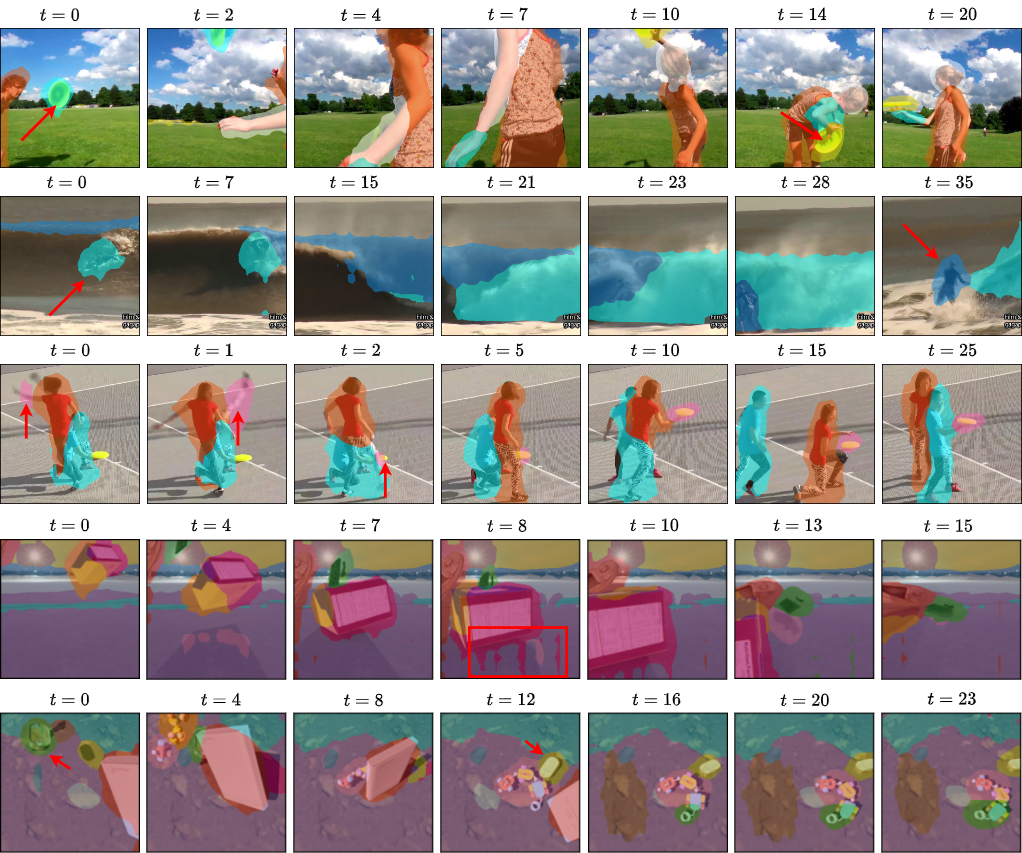}
    \caption{The visualizations depict various failure cases encountered by \method. The first three rows illustrate examples from the \method model trained on the YouTube-VIS 2021 dataset, while the last two rows are from the MOVi-C dataset. These examples highlight challenges such as failures due to complete occlusions or examples of mask ``bleeding'' artifacts. }
    \label{fig:fail_examples}
\end{figure*}

\begin{itemize}
    \item \Fig{fig:videosaur_comparison}: Comparing \method  to VideoSAUR on YouTube-VIS 2021.
    \item \Fig{fig:ablations_183}, \Fig{fig:ablations_148} and \Fig{fig:ablations_feat_rec}:   ablations of \method components.
    \item \Fig{fig:slotformer_movic}: Comparing \method and Feature Reconstruction on MOVi-C object dynamics prediction.
    \item \Fig{fig:occluded_subset}: Comparing \method  and Feature Reconstruction on MOVi-C occluded subset.
     \item \Fig{fig:movi_e}: Comparing \method to VideoSAUR on MOVi-E scene decomposition task.
    \item \Fig{fig:fail_examples}: \method failure cases. 
    
\end{itemize}

\end{document}